\documentclass[lettersize,journal]{IEEEtran}
\usepackage{color,soul} 
\soulregister\cite7 
\soulregister\citep7 
\soulregister\citet7 
\soulregister\ref7 
\soulregister\pageref7 
\usepackage{amsmath,amsfonts}
\usepackage{array}
\usepackage{booktabs}
\usepackage[caption=false,font=normalsize,labelfont=sf,textfont=sf]{subfig}
\usepackage{textcomp}
\usepackage{stfloats}
\usepackage{verbatim}
\usepackage{graphicx}
\usepackage{cite}
\usepackage{multirow}
\usepackage[pagewise]{lineno}
\usepackage[table,xcdraw]{xcolor}
\usepackage[numbers]{natbib}
\usepackage{float}
\usepackage[colorlinks=true,
            citecolor=blue,
            linkcolor=blue,
            anchorcolor=blue,
            urlcolor=blue]{hyperref}
\usepackage[linesnumbered,ruled]{algorithm2e}
\usepackage{amsthm}

\usepackage{pifont}
\usepackage{tabularray}
\usepackage{threeparttable}
\usepackage{ulem}
\usepackage{enumitem}

\begin{document}


\title{WHU-PCPR: A cross-platform heterogeneous point cloud dataset for place recognition in complex urban scenes}

\author{
\IEEEauthorblockN{
Xianghong Zou,
Jianping Li, \textit{Member}, \textit{IEEE},
Yandi Yang,
Weitong Wu,
Yuan Wang,
Qiegen Liu, \textit{Senior Member}, \textit{IEEE},
Zhen Dong, \textit{Member}, \textit{IEEE}
}

\thanks{
This study was supported by the National Natural Science Foundation Project (No. 42201477, No. 42501567), Jiangxi Provincial Natural Science Foundation (No.20252BAC200598, No.20261BCG330036, No.20242BAB21014, No.20252BAC240107). (Corresponding author: Jianping Li and Qiegen Liu)

Xianghong Zou is with the School of Advanced Manufacturing, Nanchang University, Nanchang 330031, China, and also with the State Key Laboratory of Information Engineering in Surveying, Mapping and Remote Sensing, Wuhan University, Wuhan 430079, China. (e-mail: ericxhzou@ncu.edu.cn)

Jianping Li is with the School of Electrical and Electronic Engineering, Nanyang Technological University, Singapore 639798. (e-mail: jianping.li@ntu.edu.sg)

Yandi Yang is with the Department of Geomatics Engineering, University of Calgary, Calgary T2N 1N4. (e-mail: yandi.yang@ucalgary.ca)

Weitong Wu is with the School of Earth Sciences and Engineering, Hohai University, Nanjing 211100. (e-mail: weitongwu@hhu.edu.cn)

Yuan Wang is with the School of Geography and Environment, Jiangxi Normal University, Nanchang 330022. (e-mail: wangyuanwhu@jxnu.edu.cn)

Qiegen Liu is with the School of Information Engineering, Nanchang University, Nanchang 330031, China. (e-mail: liuqiegen@ncu.edu.cn)

Zhen Dong is with the State Key Laboratory of Information Engineering in Surveying, Mapping and Remote Sensing, Wuhan University, Wuhan 430079, China. (e-mail: dongzhenwhu@whu.edu.cn)
}
}

\markboth{Journal of \LaTeX\ Class Files,~Vol.~xxx, No.~xxx, xxx~xxx}%
{Shell \MakeLowercase{\textit{et al.}}:WHU-PCPR: A cross-platform heterogeneous point cloud dataset for place recognition in complex urban scenes}

\maketitle

\begin{abstract}
Point Cloud-based Place Recognition (PCPR) demonstrates considerable potential in applications such as autonomous driving, robot localization and navigation, and map update. In practical applications, point clouds used for place recognition are often acquired from different platforms and LiDARs across varying scene. However, existing PCPR datasets lack diversity in scenes, platforms, and sensors, which limits the effective development of related research. To address this gap, we establish WHU-PCPR, a cross-platform heterogeneous point cloud dataset designed for place recognition. The dataset differentiates itself from existing datasets through its distinctive characteristics: 1) Cross-platform heterogeneous point clouds: collected from survey-grade vehicle-mounted Mobile Laser Scanning (MLS) systems and low-cost Portable helmet-mounted Laser Scanning (PLS) systems, each equipped with distinct mechanical and solid-state LiDAR sensors. 2) Complex localization scenes: encompassing real-time and long-term changes in both urban and campus road scenes. 3) Large-scale spatial coverage: featuring 82.3 km of trajectory over a 60-month period and an unrepeated route of approximately 30 km. Based on WHU-PCPR, we conduct extensive evaluation and in-depth analysis of several representative PCPR methods, and provide a concise discussion of key challenges and future research directions. The dataset and benchmark code are available at \url{https://github.com/zouxianghong/WHU-PCPR}.
\end{abstract}


\section{Introduction}\label{sec_introduction}
Place Recognition (PR) determines one’s approximate location within a pre-built map based on scene similarity, serving as an effective coarse localization technique and a fundamental component for global localization. It offers considerable application potential in fields such as autonomous driving \citep{hane20173d}, robot localization and navigation \citep{wang2021navigation,li2025ua}, multi-robot cooperative localization and mapping \citep{yuan2017cooperative,li2025graph}, map update \citep{kim2020remove}, augmented reality \citep{chi2022rebar}, and counter-terrorism rescue operations \citep{bin2017evaluation}. LiDAR (Light Detection and Ranging), as an important 3D vision sensor, is less susceptible to influences such as lighting and weather conditions compared to cameras, while providing accurate geometric, shape, and scale information, enabling Point Cloud-based Place Recognition (PCPR) to achieve stronger environmental adaptability \citep{2019NRLI,yang2024ubiquitous}.

In practical applications, point clouds used for PR are often acquired from different platforms and LiDARs \citep{zou2023patchaugnet}. Due to differences in the motion patterns, observation modes, and applicable ranges of the platforms, as well as variations in the LiDAR sensors' field of view, channel count, scanning pattern, frequency, accuracy, and measurement range, the collected point clouds exhibit significant discrepancies in accuracy, density, noise, coverage, and point distribution patterns, i.e., domain gaps \citep{qingqing2022multi,yang2024ubiquitous}. Such point clouds are thus termed heterogeneous point clouds. Taking map update as an example, a common practice is to first collect high-precision point clouds using Mobile Laser Scanning (MLS) systems as the prior map, followed by later updates with point clouds acquired by low-cost Portable Laser Scanning (PLS) systems \citep{li2023whu,LI2024228}. Before updating the map, the PLS data must be aligned to the MLS coordinate system. Global Navigation Satellite System (GNSS) provides the simplest and most direct means for this alignment. However, in GNSS-denied environments such as urban canyons, obtaining reliable positional information via GNSS becomes challenging \citep{li2026aeos}. Thus, research on PR for heterogeneous point clouds is essential.

Existing PCPR methods can be mainly categorized into two types: handcrafted feature-based and deep learning (DL)-based. The former relies on manually designed rules to extract features for PR, typically tailored to specific platforms and sensor types \citep{kim2018scan,he2016m2dp}. The latter utilizes deep neural networks to extract global features \citep{hui2021pyramid,komorowski2021minkloc3d}. These DL-based methods, though mainstream, are prone to severe overfitting on public datasets \citep{zhang2024lidar}. Besides, some reranking techniques have been introduced to enhance the PR performance \citep{zhang2022rank, vidanapathirana2023spectral}. 
Datasets play a crucial role in advancing PCPR techniques, especially for DL-based methods. However, existing PCPR datasets lack sufficient diversity in scenes, platforms, and LiDAR types to meet current research demands.
1) \textit{Limited scene diversity}: Most datasets are collected in well planned cities with short time spans and minimal scene changes. For example, Oxford Radar RobotCar dataset \citep{barnes2020oxford} is collected in the highly developed Oxford urban area within 1 month, where the roads are clean and the scene changes are minimal.
2) \textit{Limited platforms and LiDAR types}: Point clouds collected from different platforms and LiDARs exhibit significant domain gaps. While most datasets are only collected by vehicle-mounted laser scanning systems, such as KITTI Odometry \citep{geiger2012KITTI} and HeLiPR \citep{jung2024HeLiPR}. Most datasets are collected only with mechanical LiDARs, such as NCLT \citep{carlevaris2016NCLT} and Wild Place \citep{knights2023WildPlace}.
3) \textit{Limited spatial coverage}: For example, the well-established Oxford RobotCar dataset \citep{maddern20171oxford_robotcar} consists of over 100 repeated traversals along the same trajectory, which extends only approximately 10 km in length, resulting in limited spatial coverage.
Therefore, there is an urgent demand for a heterogeneous PCPR dataset that encompasses diverse scenes, platforms, and LiDARs, providing essential data support for advancing research in this field.

In this paper, we establish WHU-PCPR, a cross-platform heterogeneous point cloud dataset designed for PR. The dataset is collected across urban and campus roads in Wuhan using high-precision MLS systems and a portable PLS system, covering a total length of 82.3 km. Captured over a span of 60 months, it incorporates complex real-world scenes and diverse LiDARs, thereby reflecting practical challenges in PCPR and supporting further research. Using WHU-PCPR, we experimentally evaluate several representative PCPR methods, analyze current limitations and key challenges, and suggest potential directions for future work.

Our \textbf{main contributions} are as follows:
\begin{enumerate}[wide=0pt]
    \item A cross-platform heterogeneous PCPR dataset (WHU-PCPR) is established. It features cross-platform heterogeneous point clouds, complex localization scenes, and extensive spatial coverage, spanning a trajectory length of 82.3 km and a temporal duration of 60 months, presenting substantial challenges for PCPR.
    \item A comprehensive benchmark for PCPR is conducted on WHU-PCPR, providing researchers with an intuitive understanding of the field's development level and existing problems. Specifically, existing retrieval methods need major improvements in generalizability across scenes, platforms, and sensor types, and in robustness to viewpoint variations. Reranking techniques can effectively enhance initial retrieval results.
\end{enumerate}

The remainder of this paper is structured as follows. In section \ref{sec_related_work}, we review the existing datasets and methods for PCPR. In section \ref{sec_wuhan_pcpr}, we describe the proposed dataset WHU-PCPR. In section \ref{sec_benchmark}, we present detailed benchmark results for existing methods on WHU-PCPR. In section, \ref{sec_challeng_future_work}, we summarize the challenges and discuss possible future directions. In section \ref{sec_conclusion}, we give our conclusions.

\section{Related works}\label{sec_related_work}
\subsection{Public PCPR datasets}\label{ssec_existing_dataset}
PCPR is a highly complex task that encompasses diverse scenes, sensor configurations, and deployment platforms. In practical applications, it is essential to consider factors such as generalizability, viewpoint robustness, scene changes, computational efficiency, and storage efficiency. Furthermore, achieving efficient and reliable retrieval requires the integration of multiple technologies, including PR, reranking, and continual learning. Consequently, the advancement of these technologies necessitates the support of rich and diverse public datasets.

A range of datasets are commonly employed for PCPR research, including KITTI Odometry \citep{geiger2012KITTI}, NCLT \citep{carlevaris2016NCLT}, Oxford RobotCar \citep{maddern20171oxford_robotcar}, and so on. Among these, the Oxford RobotCar dataset is acquired by a vehicle-mounted mobile mapping system (MMS) with a SICK LMS-151 along urban roads in Oxford, with repeated traverses on the same routes. The In-house dataset \citep{uy2018pointnetvlad} is collected by a vehicle-mounted MMS with a Velodyne HDL-64E on urban roads. The KITTI Odometry dataset \citep{geiger2012KITTI}, a classic benchmark in autonomous driving, is acquired with a vehicle-mounted MMS with a Velodyne HDL-64E on urban roads in Karlsruhe. The NCLT dataset \citep{carlevaris2016NCLT} is collected by a two-wheeled robot with a Velodyne HDL-32E on a campus in Michigan, covering indoor and outdoor scenes. The ALITA dataset \citep{yin2022ALITA} is acquired by a compact unmanned vehicle with a Velodyne HDL-64E on urban and campus roads in Pittsburgh. The Apollo SouthBay dataset \citep{lu2019l3_ApolloSouthBay} is collected by a vehicle-mounted MMS with a Velodyne HDL-64E in urban, suburban, and highway areas across the San Francisco Bay Area. The Wild Place dataset \citep{knights2023WildPlace}, collected by a hand-held MMS with a Velodyne VLP-16 in forested areas of Brisbane, is the first point-cloud place recognition dataset in woodland environments. The HeLiPR dataset \citep{jung2024HeLiPR} is collected by a vehicle-mounted MMS with multiple LiDAR sensors on urban roads in Seoul, and is the first point-cloud place recognition dataset with heterogeneous LiDAR sensors.

However, a review of these datasets (detailed in Table \ref{tab:public_dataset}) reveals a common constraint: most are constructed from repeated data collection runs using the same equipment in similar urban environments. Consequently, the scene diversity and the variety of acquisition platforms and sensors are significantly limited. For example, the widely used Oxford RobotCar dataset is collected over 100 traversals using a single vehicle‑mounted LiDAR system along the same urban roads, yet its effective non‑repetitive trajectory length is only about 10 km. This homogeneity risks creating a misleading impression that current PCPR methods perform exceptionally well, masking their potential shortcomings in handling real-world complexity. To address this, this paper releases a new dataset with distinct advantages in terms of its data acquisition platforms, scene diversity, and effective spatial coverage.

\begin{table*}
\centering
\fontsize{6.5}{10}\selectfont
\caption{Commonly used public datasets for PCPR.}
\label{tab:public_dataset}
\begin{tabular}{ccccccccc} 
\hline
\textbf{Dataset}    & \textbf{Year} & \textbf{Scene}    & \textbf{Platform}     & \textbf{LiDAR}     & \textbf{Sequence}     & \begin{tabular}[c]{@{}c@{}}\textbf{Time span}\\\textbf{(month)}\end{tabular}    & \textbf{Site}      & \begin{tabular}[c]{@{}c@{}}\textbf{Length}\\\textbf{(km)}\end{tabular}  \\ 
\hline
Ford Campus \citep{pandey2011ford}                                                                & 2011          & outdoor                                                 & vehicle                                                    & Velodyne HDL-64E                                                                                          & -                                                                     & 1                                                                            & Michigan, USA       & -                                                                       \\
KITTI Odometry \citep{geiger2012KITTI}                                                             & 2012          & outdoor                                                 & vehicle                                                    & Velodyne HDL-64E                                                                                          & 22                                                                    & -                                                                            & Karlsruhe, Germany  & 39.2                                                                    \\
NCLT \citep{carlevaris2016NCLT}                                                                       & 2016          & indoor + outdoor & robot                                                      & Velodyne HDL-32E                                                                                          & 27                                                                    & 15                                                                           & Michigan, USA       & 147.4                                                                   \\
Oxford RobotCar \citep{maddern20171oxford_robotcar}                                                            & 2017          & outdoor                                                 & vehicle                                                    & SICK LMS-151                                                                                              & 133                                                                   & 19                                                                           & Oxford, UK          & 1010.5                                                                  \\
In-house \citep{uy2018pointnetvlad}                                                                   & 2018          & outdoor                                                 & vehicle                                                    & Velodyne HDL-64E                                                                                          & 15                                                                    & -                                                                            & -                   & 108.0                                                                   \\
Apollo SouthBay \citep{lu2019l3_ApolloSouthBay}                                                            & 2019          & outdoor                                                 & vehicle                                                    & Velodyne HDL-64E                                                                                          & -                                                                     & -                                                                            & San Francisco, USA  & 380.5                                                                   \\
HKUST \citep{lin2019fast}                                                                      & 2019          & outdoor                                                 & handhold                                                   & Livox Mid-40 / Mid-100                                       & -                                                                     & -                                                                            & HongKong            & -                                                                       \\
Complex Urban \citep{jeong2019complex}                                                              & 2019          &                                                         & vehicle                                                    & Velodyne VLP-16                                                                                           & 19                                                                    & -                                                                            & Korea               & 190.0\\
MulRan \citep{kim2020mulran}                                                                     & 2020          & outdoor                                                 & vehicle                                                    & Ouster OS1-64                                                                                             & 12                                                                    & 2                                                                            & Seoul, Korea        & 123.0                                                                   \\
Ford Multi-AV \citep{agarwal2020ford}                                                              & 2020          & outdoor                                                 & vehicle                                                    & Velodyne HDL-32E                                                                                          & -                                                                     & 3                                                                            & Michigan, USA       & 198.0\\
Oxford Radar RobotCar \citep{barnes2020oxford}              & 2020          & outdoor                                                 & vehicle                                                    & Velodyne HDL-32E                                                                                          & 32                                                                    & 1                                                                            & Oxford, UK          & 280.0                                                                   \\
USyd \citep{zhou2020developing}                                                                       & 2020          & outdoor                                                 & vehicle                                                    & Velodyne VLP-16                                                                                           & 52                                                                    & 13                                                                           & Sydney, Australia   & -                                                                       \\
ALITA \citep{yin2022ALITA}                                                                      & 2022          & outdoor                                                 & vehicle                                                    & Velodyne HDL-64E                                                                                          & -                                                                     & -                                                                            & Pittsburgh, USA     & 156.0                                                                   \\
KITTI-360 \citep{liao2022kitti}                                                                  & 2022          & outdoor                                                 & vehicle                                                    & Velodyne HDL-64E / SICK LMS 200                                    & 9                                                                     & -                                                                            & Karlsruhe, Germany  & 73.7                                                                    \\
HAOMO \citep{ma2022overlaptransformer}                                                                      & 2022          & outdoor                                                 & vehicle                                                    & HESAI PandarXT 32                                                                                         & 5                                                                     & 1                                                                            & Beijing, China      & -                                                                       \\
Wild Place \citep{knights2023WildPlace}                                                                 & 2023          & outdoor                                                 & handheld                                                   & Velodyne VLP-16                                                                                           & 8                                                                     & 14                                                                           & Brisbane, Australia & 32.9                                                                    \\
Boreas \citep{burnett2023boreas}                                                                     & 2023          & outdoor                                                 & vehicle                                                    & Velodyne Alpha-Prime                                                                                      & 44                                                                    & 12                                                                           & Toronto, Canada     & 350.0                                                                   \\
Pohang \citep{chung2023pohang}                                                                     & 2023          & outdoor                                                 & boat                                                       & Ouster OS1-32                                                                                             & 5                                                                     & -                                                                            & Pohang,~Korea       & 45.0                                                                    \\
HeLiPR \citep{jung2024HeLiPR}                                                                     & 2023          & outdoor                                                 & vehicle                                                    & \begin{tabular}[c]{@{}c@{}}Ouster OS2-128 / Velodyne VLP-16 \\ Livox Avia / Aeva Aeries II\end{tabular} & 19                                                                    & 53                                                                           & Seoul, Korea        & 164.0                                                                   \\
\begin{tabular}[c]{@{}c@{}}\textbf{WHU-PCPR}\\\textbf{(ours)}\end{tabular} & 2025          & outdoor                                                 & vehicle + helmet & \begin{tabular}[c]{@{}c@{}}Surveying grade LiDAR\\Livox Avia / Mid70 / Mid360\end{tabular}      & 6                                                                     & 60                                                                           & Wuhan, China        & 82.3                                                                    \\
\hline
\end{tabular}
\end{table*}

\subsection{PCPR methods}\label{ssec_existing_method}
Existing PCPR methods include two steps: retrieval and reranking \citep{vidanapathirana2023spectral}. The former performs initial PR using either handcrafted features or deep learning-based features. The latter employs techniques like query expansion \citep{chum2007total} and point cloud registration \citep{zhang2022rank} to refine initial retrieval results through additional information, and it's an optional step.

\textbf{Retrieval}. These methods can be categorized into handcrafted feature-based and learning-based. \textit{Handcrafted feature-based} methods extract features for PR based on manual rules, including projection, histogram, and so on. Projection-based methods convert point clouds into 2D images, such as bird's eye view (BEV) images \citep{kim2018scan,liang2020novel}. Histogram-based methods divide the point cloud into grids to build statistical histograms \citep{cop2018delight,lin2019fast}. Other methods directly extract features from point clouds \citep{bosse2013place,dube2017segmatch}. These methods rely on manual rules, typically only applicable to specific platforms and LiDARs, and are sensitive to viewpoint variations, occlusions, and noises.

\textit{Learning-based} methods utilize deep networks to extract global features, and can be divided into three categories: point-based \citep{uy2018pointnetvlad,zou2023patchaugnet,xie2024look}, voxel-based \citep{komorowski2021minkloc3d,vidanapathirana2022logg3d}, and projection-based \citep{chen2021overlapnet,luo2023bevplace}. Point-based and Voxel-based methods extract global features from raw and voxelized point clouds respectively. Projection-based method extracts features from 2D projection images. These methods require large-scale geotagged data for training. Their transferability across diverse scenes and LiDARs, as well as robustness to scene changes, remain key challenges for improvement.

\textbf{Reranking}. Single retrieval consistently struggles in challenging scenes. It's necessary to adjust the initial retrieval results using additional information, i.e. reranking. Reranking methods for PCPR can be categorized into two types: query expansion and geometric consistency. Query expansion utilizes the features of the retrieved top-k items to update the feature of query item for reranking \citep{chum2007total,radenovic2018fine}. Methods based on geometric consistency primarily leverage registration scores to rerank the top-k items \citep{zhang2022rank,vidanapathirana2023spectral,zhang2025Enhancing}. Existing reranking methods suffer from both noise sensitivity and computational efficiency. Among them, methods based on geometric consistency tend to be more reliable.

\section{Construction of WHU-PCPR dataset}\label{sec_wuhan_pcpr}
\subsection{Data collection}\label{ssec_data_collection}
\begin{figure*}
    \centering
    \includegraphics[width=1.0\linewidth]{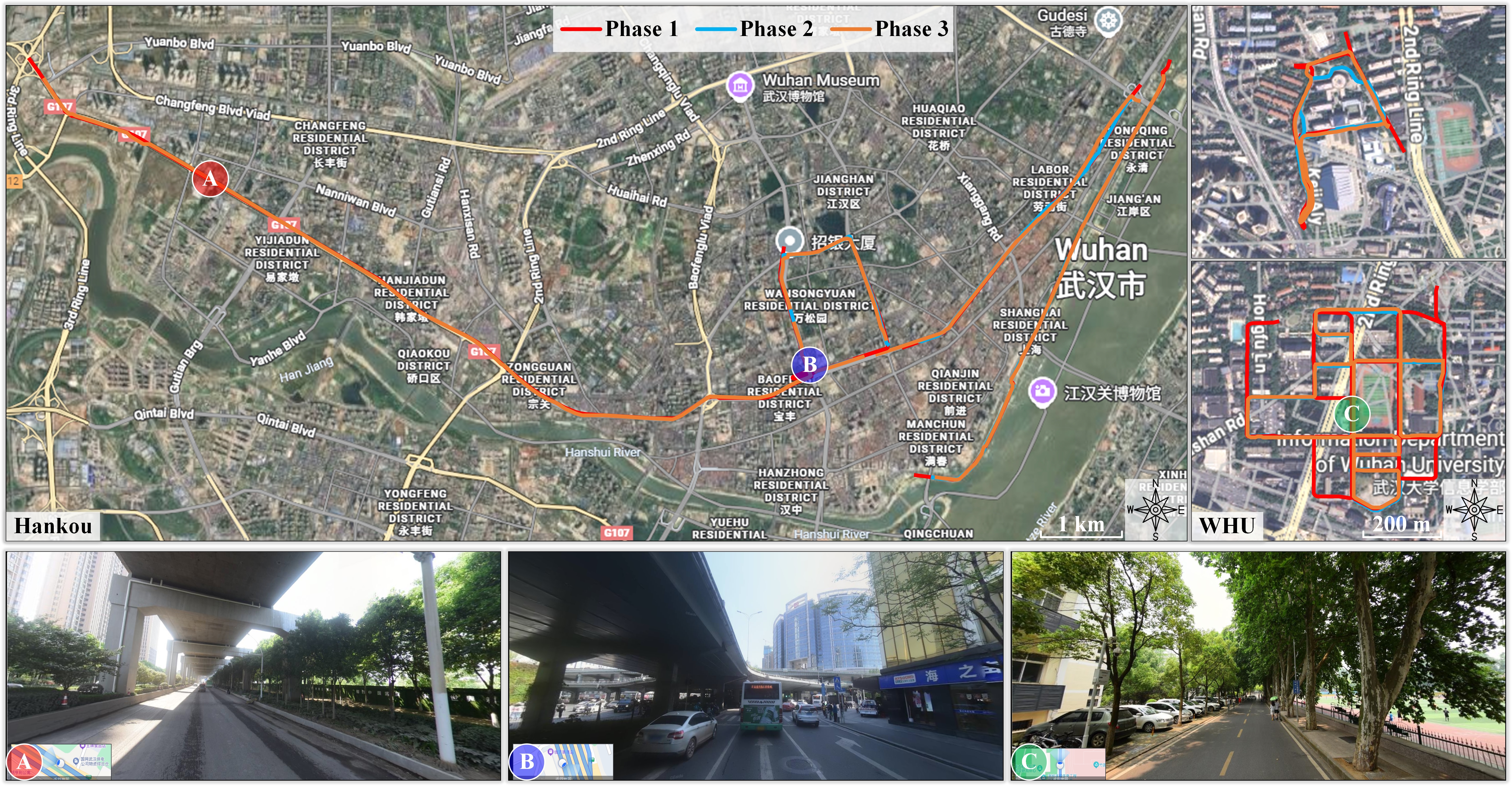}
    \caption{Overview of WHU-PCPR dataset. A, B, and C represent three typical scenes.}
    \label{fig:WHU-PCPR-overview}
\end{figure*}

WHU-PCPR dataset is collected in the urban area of Wuhan, China using high-precision MLS systems and portable PLS systems, as shown in Fig. \ref{fig:WHU-PCPR-overview}. Details are as follows:

\textbf{Acquisition area.} 
WHU-PCPR comprises two distinct regions: Hankou and WHU, as shown in Fig. \ref{fig:WHU-PCPR-overview}. Hankou data is collected along typical urban thoroughfares in Wuhan, including Jiefang Road, Yanjiang Road, and Zhongshan Park. The scenes feature dense high-rise buildings, overpasses, street trees, and dynamic objects. WHU data covers areas within Wuhan University, including computer science (CS) college and Information (Info) campus. These scenes are characterized by confined spaces, dense vegetation, and highly unstructured terrain. Over the 60-month period, some areas in both Hankou and WHU show significant changes.

\textbf{Acquisition equipments.}
WHU-PCPR collects heterogeneous point clouds with survey-grade MLS systems and low-cost portable PLS systems, as shown in Fig. \ref{fig:wuhan_pcpr_characteristics}. The MLS systems (Hi-Target HiScan-VUX and CHCNAV Alpha 3D) are equipped with high-precision positioning and orientation systems as well as single-beam, high-frequency mechanical LiDARs, and the collected point clouds are of high precision, high density, and wide coverage. The PLS system, designated WHU-Helmet \citep{li2023whu}, is equipped with a semi-solid state LiDAR, i.e. Livox Mid70, Avia or Mid360. Solid-state LiDARs use non-repetitive scanning and typically have a shorter range, resulting in point clouds with lower accuracy, lower density, and reduced coverage.

\textbf{Statistics of WHU-PCPR.}
WHU-PCPR comprises 6 sequences spanning a total trajectory length of 82.3 km over 60 months. Spatially, the dataset comprises 38,616 submaps, with 31,466 from MLS and 7,150 from PLS. Correspondingly, it covers 82.3 km of trajectories, with 30.8 km from MLS and 51.5 km from PLS. Temporally, the Hankou subset spans 60 months, while the WHU subset covers 20 months. A detailed summary is provided in Table \ref{tab:WHU-PCPR}.

\begin{table*}
\centering
\fontsize{6.4}{10}\selectfont
\caption{Detailed description of WHU-PCPR dataset.}
\label{tab:WHU-PCPR}
\begin{tblr}{
  cells = {c},
  cell{2}{1} = {r=3}{},
  cell{2}{3} = {r=3}{},
  cell{5}{1} = {r=3}{},
  cell{5}{3} = {r=3}{},
  hline{1-2,5,8} = {-}{},
}
Region & Phase & Scene       & Platform & LiDAR                & Acquisition time                   & Length (km) & {Submap (no dynamics):\\Train/Test} & {Submap (with dynamics):\\Train/Test} \\
Hankou & 1     & Urban road  & Vehicle  & Hi-Target HiScan-VUX & 2019-12-11, 2020-03-20, 2020-03-22 & 24.6       & 10329 / 1563                        & 10337/1567                            \\
       & 2     &             & Helmet   & Livox Avia           & 2022-08-23                         & 15.4       & 5467 / 1594                         & 5521/1617                             \\
       & 3     &             & Helmet   & Livox Mid360         & 2023-07-21, 2023-07-23, 2024-12-13 & 27.5       & 0 / 12513                           & 0/13124                               \\
WHU    & 1     & Campus road & Vehicle  & CHCNAV Alpha3D       & 2021-11-20                         & 6.2        & 0 / 2995                            & 0/3092                                \\
       & 2     &             & Helmet   & Livox Avia / Mid70   & 2021-11-12, 2022-07-29             & 4.1        & 0 / 2023                            & 0/1969                                \\
       & 3     &             & Helmet   & Livox Mid360         & 2023-07-18                         & 4.5        & 0 / 2132                            & 0/2176                                
\end{tblr}
\begin{tablenotes}
\footnotesize
\item \textit{Note: when removing dynamic objects, submaps with too few points are discarded, yielding a lower submap count than retaining them.}
\end{tablenotes}
\end{table*}

\subsection{Data preprocessing}\label{ssec_data_preprocess}

\begin{figure*}
    \centering
    \includegraphics[width=1\linewidth]{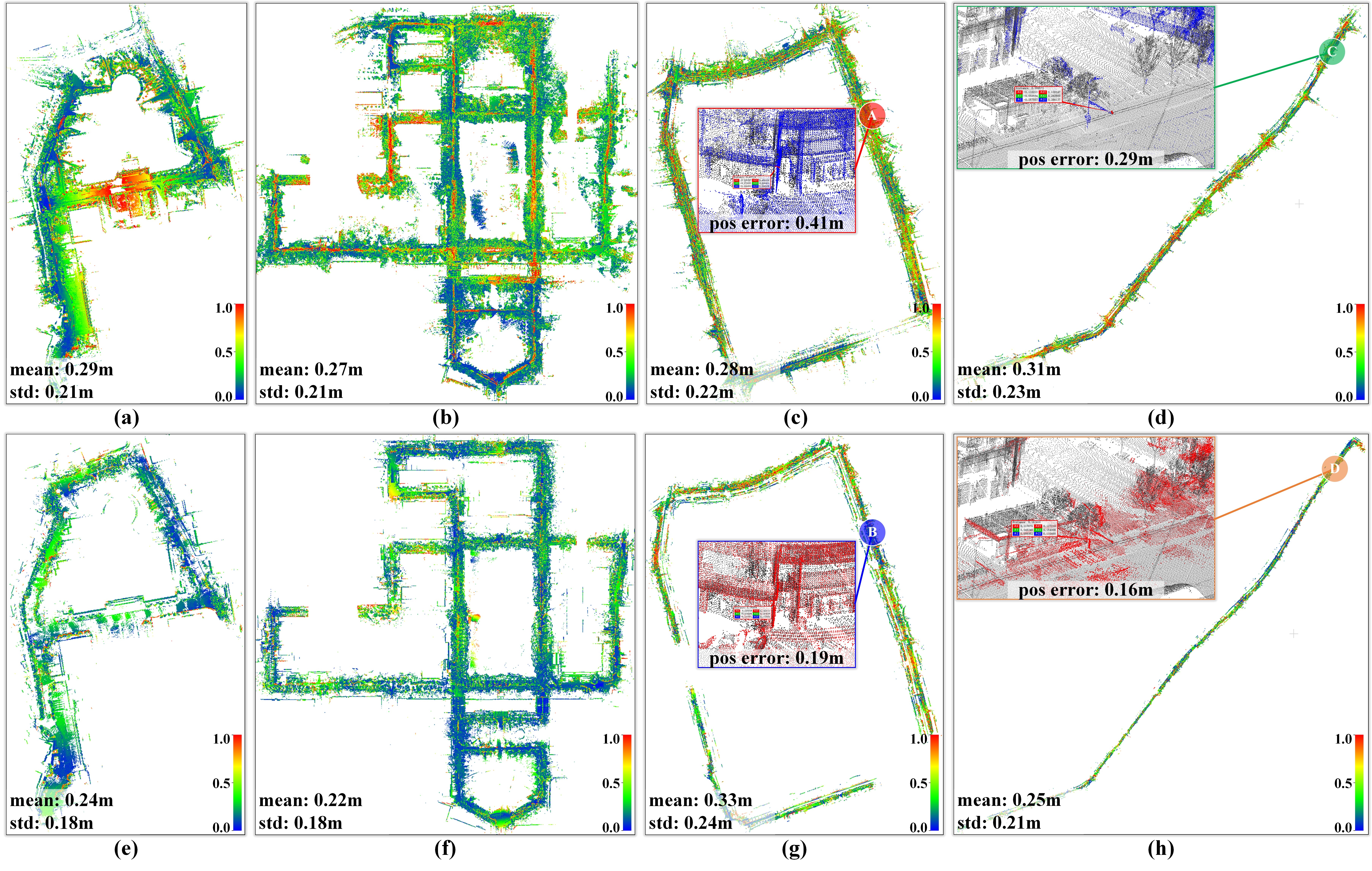}
    \caption{Cloud to cloud distance in WHU-PCPR. (a) WHU 1\&2 (CS College),  (b) WHU 1\&2 (Info Campus), (c) Hankou 1\&2 (Zhongshan Park), (d) Hankou 1\&2 (Jiefang Road 1), (e) WHU 1\&3 (CS College), (f) WHU 1\&3 (Info Campus), (g) Hankou 1\&3 (Zhongshan Park), (h) Hankou 1\&3 (Jiefang Road 1). A, B, C, and D are the positional errors of manually selected corresponding points (gray/blue/red: phase 1/2/3).}
    \label{fig:cloud_distance}
\end{figure*}

\textbf{Data alignment.}
The MLS trajectories, obtained from high-precision GNSS and IMU with a unified coordinate system, serves as the reference for data alignment. The initial PLS trajectory, derived using Fast-LIO2 \citep{2022FAST}, is locally smooth but suffers from significant cumulative drift. Therefore, we manually register the PLS data to the MLS data. Specifically, we segment the PLS trajectory and apply the point cloud registration tool of CloudCompare\footnote{https://www.cloudcompare.org/} to align both the segmented PLS point clouds and trajectory to the MLS coordinate system by manually selecting corresponding point pairs. The average registration accuracy of the multi-phase point clouds is 0.28 m, which satisfies the experimental requirements.
More details refer to Fig. \ref{fig:cloud_distance}.

\textbf{Submap construction.}
Circular submaps with a 30-meter radius are generated at 2-meter intervals along each trajectory. As the full high‑density point cloud data is made available, users may subsequently modify the submap radius as needed. Before submap construction, dynamic objects are removed from the PLS point clouds using Octomap \citep{2013OctoMap}. Each submap is generated by accumulating point clouds from several scans collected over a 2‑meter segment along the trajectory, with ground points removed via cloth simulation filtering \citep{2016CSF}.

\textbf{Data splitting.}
The dataset is partitioned using rectangular regions along the x/y axes. For sequences Hankou 1\&2, submaps within specific 120mx120m squares are designated as the test set, while the rest form the training set, resulting in an approximate 5:1 train-test ratio. All submaps from the remaining sequences are used for testing. In the training set, submaps are defined as positive pairs when their centers are within 15 m, and as negative pairs when the distance exceeds 60 m. During testing, a retrieval is considered successful if the retrieved submap is within 30 m of the query.

\begin{figure}
    \centering
    \includegraphics[width=1\linewidth]{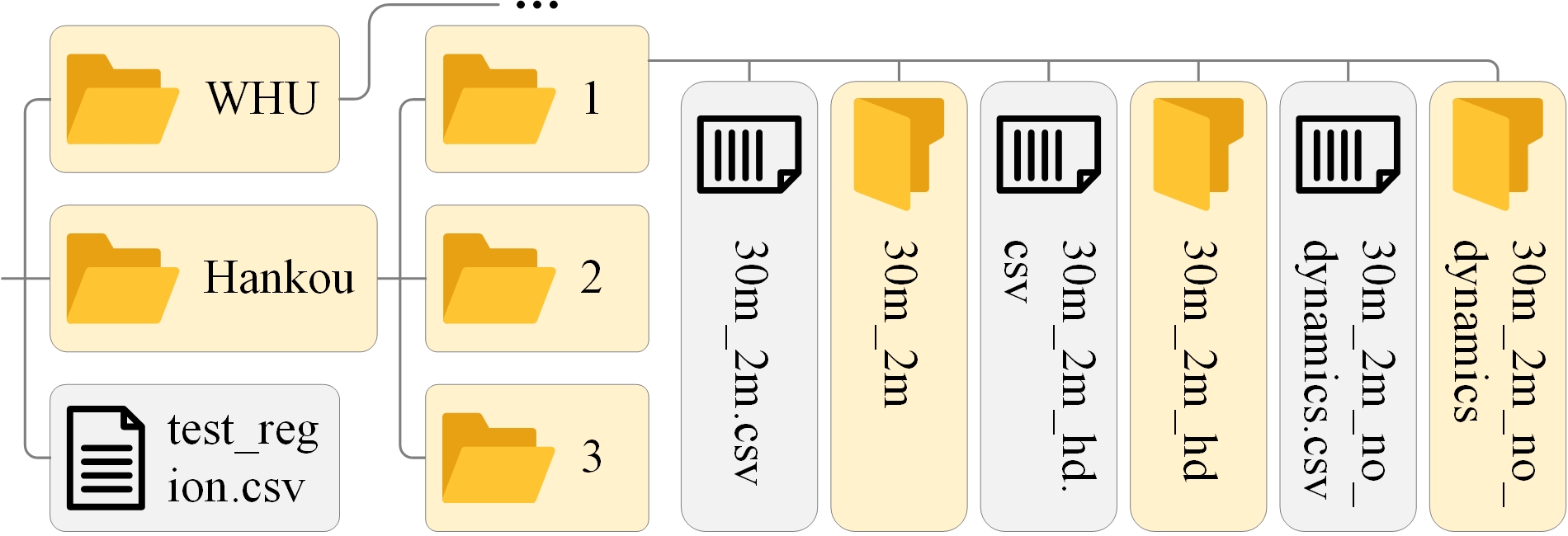}
    \caption{File structure of the dataset.}
    \label{fig:WHU-PCPR-data-format}
\end{figure}

\textbf{Data format.}
Fig. \ref{fig:WHU-PCPR-data-format} presents the file structure of WHU-PCPR. The dataset consists of two scenes. Each scene has three sequences, and each sequence has three kinds of submap construction approaches. Taking sequence 1 of Hankou as an example, it provides three kinds of submaps. "30m\_2m" denotes submaps that include dynamic objects; "30m\_2m\_hd" refers to its corresponding high-density point cloud submaps; "30m\_2m\_no\_dynamics" indicates submaps without dynamic objects. "30m\_2m" and "30m\_2m\_no\_dynamics" folders contain submap files in ".bin" format that can be loaded using NumPy. "30m\_2m\_hd" folders contain submap files in ".las" format. Besides, "test\_region.csv" provides the center coordinates for the square test regions.

\subsection{Characteristics of WHU-PCPR}\label{ssec_data_advantage}
\begin{figure*}
    \centering
    \includegraphics[width=1\linewidth]{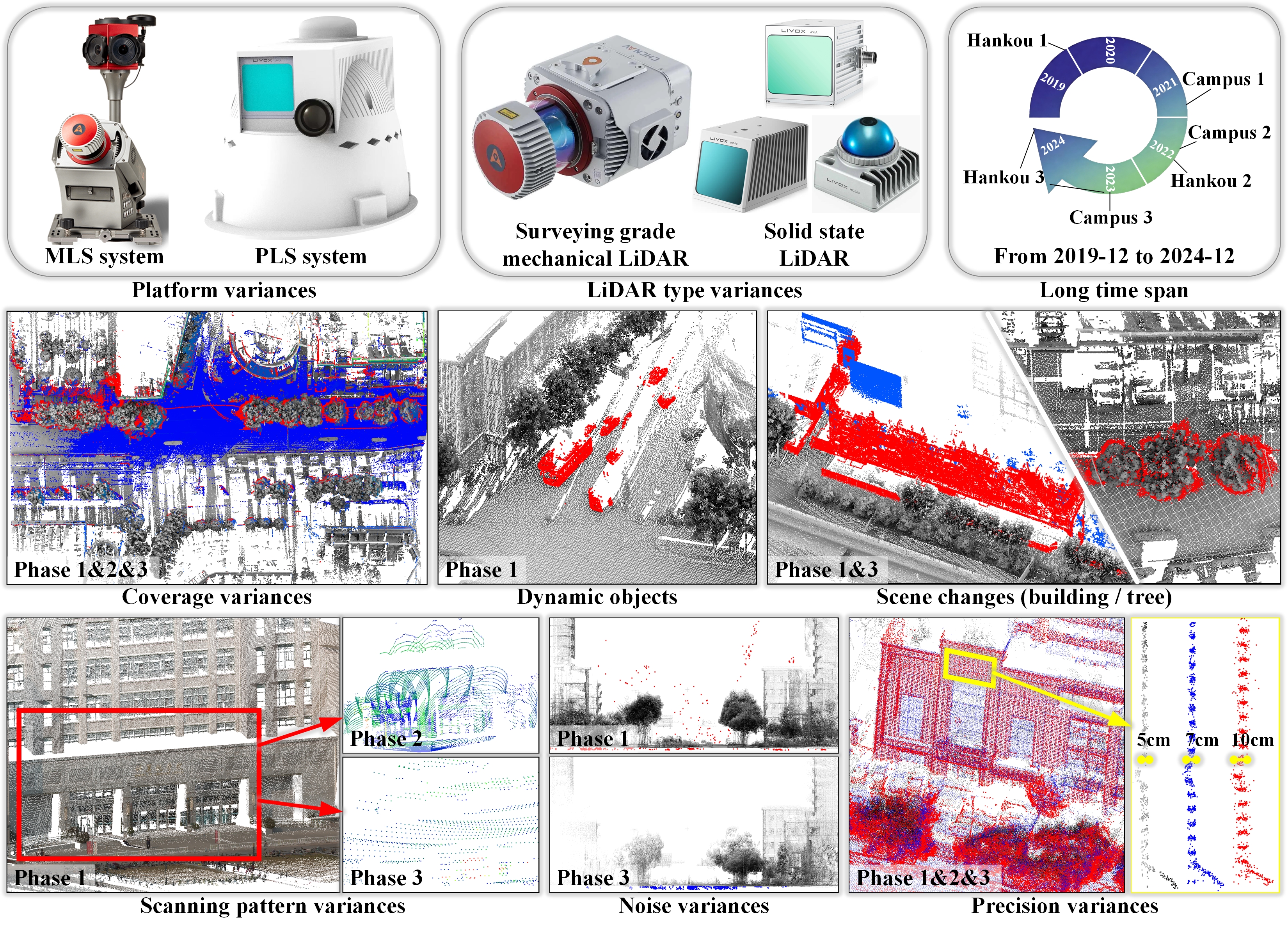}
    \caption{Characteristics of WHU-PCPR.}
    \label{fig:wuhan_pcpr_characteristics}
\end{figure*}

As shown in Fig. \ref{fig:wuhan_pcpr_characteristics}, the WHU-PCPR dataset exhibits three key characteristics:
\begin{enumerate}[wide=0pt]
    \item \textbf{Cross-platform heterogeneous point clouds}:
First, the dataset is collected from two distinct platforms. The motion ranges and patterns of these platforms differ significantly. Second, different platforms often carry different types of LiDARs, varying in terms of measurement range, accuracy, scanning frequency, and scanning pattern. Third, The MLS system incorporates high‑precision GNSS and IMU sensors, while the helmet‑mounted PLS system relies solely on low‑cost GNSS and IMU sensors, resulting in a considerable difference in their data acquisition accuracy. Due to the above reasons, the collected point clouds exhibit significant differences in accuracy, density, coverage, and noise levels.
    \item \textbf{Complex localization scenes}:
First, campus roads typically exhibit dense vegetation, few artificial structures, and highly unstructured surroundings, while urban roads are marked by abundant street trees, tall buildings, overpasses, and dynamic objects like vehicles and pedestrians. Second, the three-phase data collection spans a five-year period, during which certain scenes have undergone substantial changes, such as buildings and trees.
    \item \textbf{Large-scale spatial coverage}:
In contrast to the widely recognized Oxford RobotCar dataset, which has an unrepeated route of about 10 km, the WHU-PCPR dataset covers roughly 30 km, representing a threefold increase in spatial coverage. Besides, MLS point clouds are collected on the road, while PLS point clouds are collected roadside. The positional offset between them presents a challenge for PR. For related discussion, see Scan Context++ \citep{kim2021scan}.
\end{enumerate}

\section{Benchmark}\label{sec_benchmark}
We evaluate representative PCPR methods on WHU-PCPR to illustrate the challenges it presents and expose the limitations of existing research. Specically, we first train retrieval baselines on training sets (Hankou 1 and Hankou 2, i.e. Hankou 1\&2) of WHU-PCPR, then perform retrieval and reranking on testing sets of WHU-PCPR and Oxford RobotCar (denoted as Oxford) to evaluate the baselines and analyze their sensitivity to viewpoint variations. Experiments on WHU-PCPR have two protocols. In protocol 1, all retrieval networks are trained and tested on submaps without dynamic objects, and all re‑ranking methods are tested on the same submaps. In protocol 2, all retrieval networks are trained and tested on submaps containing dynamic objects, and all re‑ranking methods are tested on the same submaps. We primarily employ protocol 1, supplemented by protocol 2.

\subsection{Experiment setting}\label{ssec_exp_setting}
\textbf{Baselines.}
For retrieval, we select five representative methods as baselines. PointNetVLAD \citep{uy2018pointnetvlad} is the first DL-based PCPR method. PPTNet \citep{hui2021pyramid} is a point-based state-of-the-art (SOTA) method using Transformer and NetVLAD. MinkLoc3D \citep{komorowski2021minkloc3d} is the first voxel-based method using sparse convolution. EgoNN \citep{komorowski2021egonn} extracts local feature for 6DoF relocalization on the basis of MinkLoc3D. LoGG3D-Net \citep{vidanapathirana2022logg3d} is a SOTA method using sparse point-voxel convolution.
For reranking, we select three representative methods as baselines. AlphaQE \citep{radenovic2018fine} (denoted as aQE) is a classic query expansion method. Rank-PointRetrieval \citep{zhang2022rank} (denoted as RPR) is typical reranking method based on geometric consistency. SpectralGV \citep{vidanapathirana2023spectral} (denoted as SGV) is SOTA reranking method for PCPR.

\textbf{Evaluation metrics.}
We employ $Recall@topN$ (denoted as $R@N$) and $Precision@topN$ (denoted as $P@N$) as quantitative metrics to evaluate PCPR performance. $R@N$ measures the ratio of successfully retrieved submaps, while $P@N$ indicates the proportion of correct results among the retrieved top $N$ submaps. For retrieval, we report only $R@N$, especically $R@1$ and $R@1\%$. For reranking, we calculate both $R@N$ and $P@N$.

\textbf{Implementation details.}
All submaps are downsampled to 4096 points using farthest point sampling, and their coordinates are normalized to the range [-1, 1]. We primarily adopt the official implementations of all baseline methods. For retrieval, following the practice of MinkLoc3D \citep{komorowski2021minkloc3d}, we supervise all networks using the batch hard triplet margin loss and do not use data augmentation. For reranking, each point cloud submap retains 128 local features. For better performance on heterogeneous point clouds, we normalize the global features with reference to PatchAugNet \citep{zou2023patchaugnet}. All retrieval baselines are implemented in PyTorch and trained on an NVIDIA® RTX 4080s GPU. All reranking baselines run on Intel® Core i9-14900KF CPU.

\subsection{Quantitative evaluation for PCPR}\label{ssec_eval_PCPR}
\subsubsection{Retrieval results}\label{sssec_eval_PCPR_retrieval}
\begin{figure*}
    \centering
    \includegraphics[width=1\linewidth]{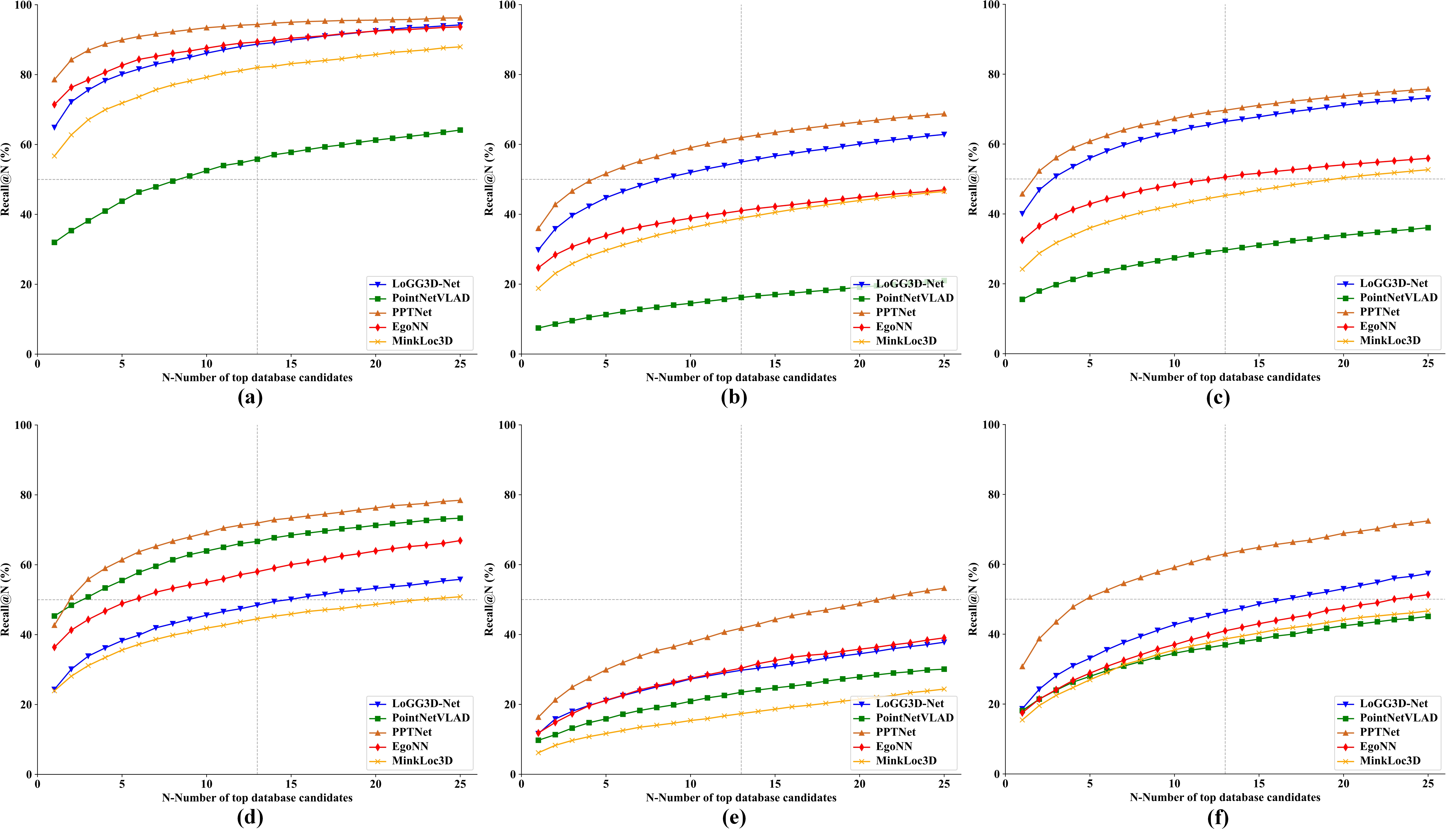}
    \caption{Recall curves of retrieval baselines on WHU-PCPR with protocol 1. (a) Hankou 1\&2, (b) Hankou 1\&3, (c) Hankou 2\&3, (d) WHU 1\&2, (e) WHU 1\&3, (f) WHU 2\&3.}
    \label{fig:PR_recall_curve_no_dynamics}
\end{figure*}

Table \ref{tab:pr_baselines_no_dynamics}, Table \ref{tab:pr_baselines_with_dynamics}, and Fig. \ref{fig:PR_recall_curve_no_dynamics} present the retrieval results of various baselines. Overall, LoGG3D-Net, a retrieval method utilizing sparse point-voxel convolution, achieves the best performance. We analyze these evaluation results from the perspectives of generalization, viewpoint robustness, as well as efficiency.

\textbf{Generalization.}
As shown in Table \ref{tab:pr_baselines_no_dynamics} and Fig. \ref{fig:PR_recall_curve_no_dynamics} (a, d), after being trained on the Hankou data from phase 1 and 2 (denoted as Hankou 1\&2), the retrieval methods perform the best on Hankou 1\&2, followed by WHU 1\&2. Taking LoGG3D-Net as an example, it attains an $R@1$ of 80.70\% and $R@1\%$ of 97.05\% on Hankou 1\&2. However, its performance drops significantly on WHU 1\&2, with $R@1$ and $R@1\%$ falling to 40.13\% and 74.71\%, respectively. This decline stems from substantial differences in the shape, quantity, and distribution of surface features across scenes. Hankou road scenes are typically more open, featuring numerous high-rise buildings, elevated structures, and street trees. In contrast, WHU road scenes are characterized by dense vegetation and fewer artificial objects (see Fig. \ref{fig:WHU-PCPR-overview}). However, existing methods are significantly overfitting on Hankou data, limiting their generalizability to other scenes.

As shown in Table \ref{tab:pr_baselines_no_dynamics} and Fig. \ref{fig:PR_recall_curve_no_dynamics} (a-c), after being trained on Hankou 1\&2, the retrieval methods also exhibit a significant performance drop on Hankou 1\&3 and Hankou 2\&3. For example, $R@1$ and $R@1\%$ on Hankou 1\&3 are only 36.25\% and 70.25\%, respectively, while they are 46.55\% and 77.74\% on Hankou 2\&3. 

This decline is attributed to the domain gap arising from different data acquisition systems: Phase 1 data is collected using survey-grade MLS systems, whereas phases 2 and 3 employ low-cost PLS systems with different LiDARs. This results in heterogeneous point clouds with distinct characteristics (see Fig. \ref{fig:wuhan_pcpr_characteristics}). Consequently, the methods overfit to the specific acquisition equipments of the training data, limiting their generalizability to data from other acquisition equipments.

As shown in Table \ref{tab:pr_baselines_no_dynamics} and Fig. \ref{fig:PR_recall_curve_no_dynamics} (e-f), all methods face challenges from both the differences in scenes and equipments, leading to a further performance decline. Taking LoGG3D-Net as an example, $R@1$ and $R@1\%$ on WHU 1\&3 are only 18.85\% and 51.79\%, while they are 28.36\% and 68.12\% on WHU 2\&3. The superior performance on WHU 2\&3 over WHU 1\&3 indicates that the domain gap between mechanical LiDAR and solid-state LiDAR is greater than that between different solid-state LiDARs themselves. Furthermore, despite significant differences in scenes and acquisition equipments between Oxford and Hankou 1\&2, LoGG3D-Net performs considerably better on Oxford than on WHU 1\&2. The reason is that there is no domain gaps between query and map point clouds within Oxford itself.

\begin{figure*}
    \centering
    \includegraphics[width=0.7\linewidth]{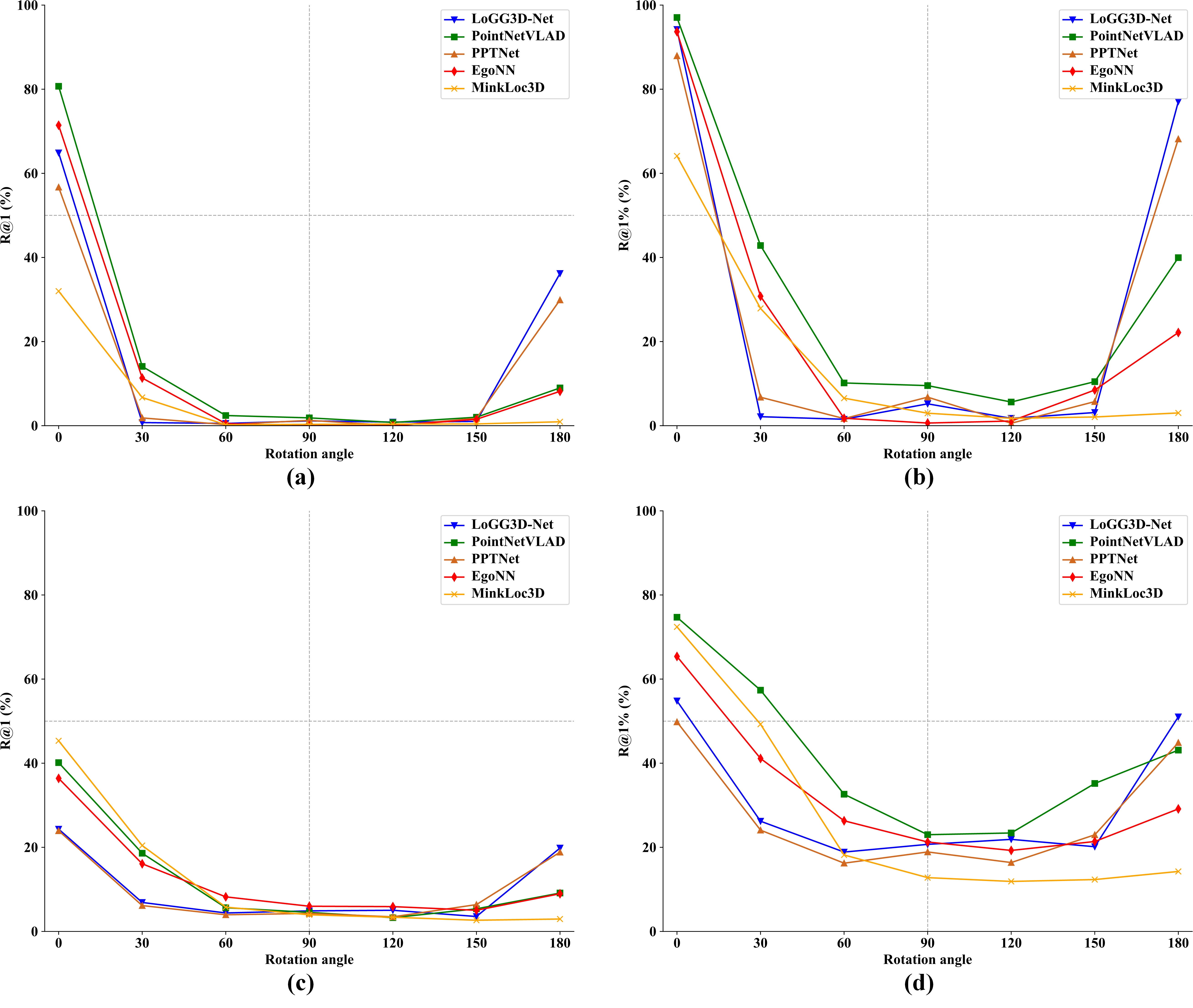}
    \caption{Retrieval results of various baselines with different viewpoints on WHU-PCPR with protocol 1. (a) $R@1$ on Hankou 1\&2, (b) $R@1\%$ on Hankou 1\&2, (c) $R@1$ on WHU 1\&2, (d) $R@1\%$ on WHU 1\&2.}
    \label{fig:Viewpoint_recall_curve_no_dynamics}
\end{figure*}

\textbf{Robustness to viewpoint.}
Fig. \ref{fig:Viewpoint_recall_curve_no_dynamics} presents the PR performance of various retrieval baselines under different viewpoints. As shown, point-based and voxel-based methods like PointNetVLAD and MinkLoc3D exhibit high sensitivity to viewpoint variations. When query submaps are rotated 30 degrees around the z-axis, $R@1$ and $R@1\%$ of all methods drop below 20\%. When rotating 90 degrees around the z-axis, all methods are basically unusable. The experimental results demonstrate that viewpoint variation poses a significant challeng to PCPR tasks, and this problem cannot be solved or alleviated by reranking.

\textbf{Robustness to dynamics.}
Table \ref{tab:pr_baselines_with_dynamics} presents the PR performance of various retrieval baselines in protocol 2. By comparing with Table \ref{tab:pr_baselines_no_dynamics}, It can be observed that LoGG3D-Net consistently achieves the best performance, with its average recall under protocol 2 improving by 5.03\% to 8.98\% relative to that under protocol 1. This improvement can be attributed to two primary reasons. Firstly, submaps in protocol 2 retain all dynamic objects, where dynamic entities such as vehicles and pedestrians serve as negative anchors that compel the model to attend to stable static backgrounds for discrimination, thereby improving feature discriminability. Secondly, despite utilizing Octomap to remove dynamic points, submaps in protocol 1 still contain some dynamic objects that are not successfully eliminated, thereby posing a greater challenge to the model trained under protocol 1.

\textbf{Efficiency.}
As shown in Table \ref{tab:pr_baselines_with_dynamics_efficiency}, all methods demonstrate high computational efficiency, with each retrieval time not exceeding 20 ms. Overall, voxel-based methods are less efficient than point-based ones, and LoGG3D-Net, which uses sparse point-voxel convolution, has the highest computational cost.

In summary, the diversity in scenes, collection platforms, and LiDAR types within WHU-PCPR presents substantial challenges for PCPR, and there is significant room for improvement in the performance of all retrieval baselines. Thus, WHU-PCPR is valuable for advancing research toward robust PCPR methods that generalize across varying scenes, platforms, and LiDAR types.

\begin{table*}
\centering
\fontsize{6.4}{10}\selectfont
\caption{Retrieval results on WHU-PCPR and Oxford with protocol 1.}
\label{tab:pr_baselines_no_dynamics}
\begin{tabular}{ccccccccccccccccc} 
\hline
\multirow{2}{*}{Method} & \multicolumn{2}{c}{Hankou 1\&2} & \multicolumn{2}{c}{Hankou 1\&3} & \multicolumn{2}{c}{Hankou 2\&3} & \multicolumn{2}{c}{WHU 1\&2} & \multicolumn{2}{c}{WHU 1\&3} & \multicolumn{2}{c}{WHU 2\&3} & \multicolumn{2}{c}{Oxford}      & \multicolumn{2}{c}{Average}      \\ 
\cline{2-17}
                        & R@1            & R@1\%                        & R@1            & R@1\%                        & R@1            & R@1\%                        & R@1            & R@1\%                     & R@1            & R@1\%                     & R@1            & R@1\%                     & R@1            & R@1\%          & R@1            & R@1\%           \\ 
\hline
PointNetVLAD            & 32.43          & 64.01                        & 7.45           & 20.94                        & 15.81          & 36.04                        & \textbf{45.67} & \uline{72.76}             & 9.96           & 29.51                     & 18.33          & 42.99                     & \textbf{74.45} & \textbf{88.92} & 29.16          & 50.74           \\
PPT-Net                 & 56.71          & 87.95                        & 18.81          & 46.54                        & 24.20          & 52.66                        & 23.94          & 49.82                     & 6.17           & 22.84                     & 15.46          & 44.38                     & 65.94          & 86.78          & 30.18          & 55.85           \\
MinkLoc3D               & 64.87          & \uline{94.22}                & \uline{29.82}  & \uline{62.85}                & \uline{40.05}  & \uline{73.18}                & 24.32          & 54.81                     & 11.63          & 36.26                     & \uline{18.66}  & \uline{53.69}             & 60.36          & 84.32          & 35.67          & \uline{65.62}   \\
EgoNN                   & \uline{71.43}  & 93.68                        & 24.69          & 47.00                        & 32.50          & 55.92                        & 36.39          & 65.38                     & \uline{11.82}  & \uline{37.49}             & 17.51          & 48.01                     & 62.50          & 84.04          & \uline{36.69}  & 61.65           \\
LoGG3D-Net              & \textbf{80.70} & \textbf{97.05}               & \textbf{36.25} & \textbf{70.25}               & \textbf{46.55} & \textbf{77.74}               & \uline{40.13}  & \textbf{74.71}            & \textbf{18.85} & \textbf{51.79}            & \textbf{28.36} & \textbf{68.12}            & \uline{68.50}  & \uline{88.20}  & \textbf{45.62} & \textbf{75.41}  \\
\hline
\end{tabular}
\end{table*}

\begin{table*}
\centering
\fontsize{6.4}{10}\selectfont
\caption{Retrieval results on WHU-PCPR and Oxford with protocol 2.}
\label{tab:pr_baselines_with_dynamics}
\begin{tabular}{ccccccccccccccccc}
\hline
\multirow{2}{*}{Method} & \multicolumn{2}{c}{Hankou 1\&2}    & \multicolumn{2}{c}{Hankou 1\&3}    & \multicolumn{2}{c}{Hankou 2\&3}    & \multicolumn{2}{c}{WHU 1\&2}    & \multicolumn{2}{c}{WHU 1\&3}    & \multicolumn{2}{c}{WHU 2\&3}    & \multicolumn{2}{c}{Oxford}      & \multicolumn{2}{c}{Average}      \\
                        & R@1            & R@1\%          & R@1            & R@1\%          & R@1            & R@1\%          & R@1            & R@1\%          & R@1            & R@1\%          & R@1            & R@1\%          & R@1            & R@1\%          & R@1            & R@1\%           \\ 
\hline
PointNetVLAD            & 25.37          & 57.13          & 19.60          & 40.37          & 33.97          & 61.47          & \textbf{45.51} & \uline{69.67}  & \textbf{45.76} & \textbf{69.84} & \uline{50.03}  & 75.89          & \textbf{74.22} & \textbf{88.74} & 42.07          & 66.16           \\
PPT-Net                 & 54.05          & \uline{90.73}  & 36.83          & 72.32          & 50.58          & \uline{86.62}  & 25.05          & 52.79          & 20.65          & 45.97          & 40.97          & 77.65          & \uline{67.60}  & \uline{87.60}  & \uline{42.25}  & \uline{73.38}   \\
Minkloc3D               & \uline{58.51}  & \textbf{92.96} & \uline{41.61}  & \textbf{75.39} & \uline{55.68}  & 86.36          & 18.26          & 46.00          & 18.14          & 44.44          & 42.23          & \uline{79.92}  & 48.13          & 74.78          & 40.37          & 71.41           \\
EgoNN                   & 48.45          & 82.34          & 30.22          & 57.51          & 49.59          & 79.93          & 34.47          & 69.27          & 34.45          & 65.30          & 38.70          & 71.13          & 41.51          & 70.05          & 39.63          & 70.79           \\
LoGG3D-Net              & \textbf{63.97} & 89.35          & \textbf{45.20} & \uline{74.71}  & \textbf{64.89} & \textbf{89.71} & 43.27          & \textbf{71.42} & \uline{41.64}  & \uline{67.98}  & \textbf{59.26} & \textbf{85.35} & 63.95          & 84.59          & \textbf{54.60} & \textbf{80.44}  \\
\hline
\end{tabular}
\end{table*}

\begin{table}
\centering
\fontsize{6.4}{10}\selectfont
\caption{Retrieval efficiency of PR baseline methods.}
\label{tab:pr_baselines_with_dynamics_efficiency}
\begin{tabular}{cccccc}
\hline
Method    & PointNetVLAD  & PPT-Net & MinkLoc3D    & EgoNN & LoGG3D-Net  \\ 
\hline
time (ms) & \textbf{2.96} & 14.72   & \uline{4.72} & 16.29 & 19.13       \\
\hline
\end{tabular}
\end{table}

\subsubsection{Reranking results}\label{sssec_eval_PCPR_rerank}
\begin{figure*}
    \centering
    \includegraphics[width=1\linewidth]{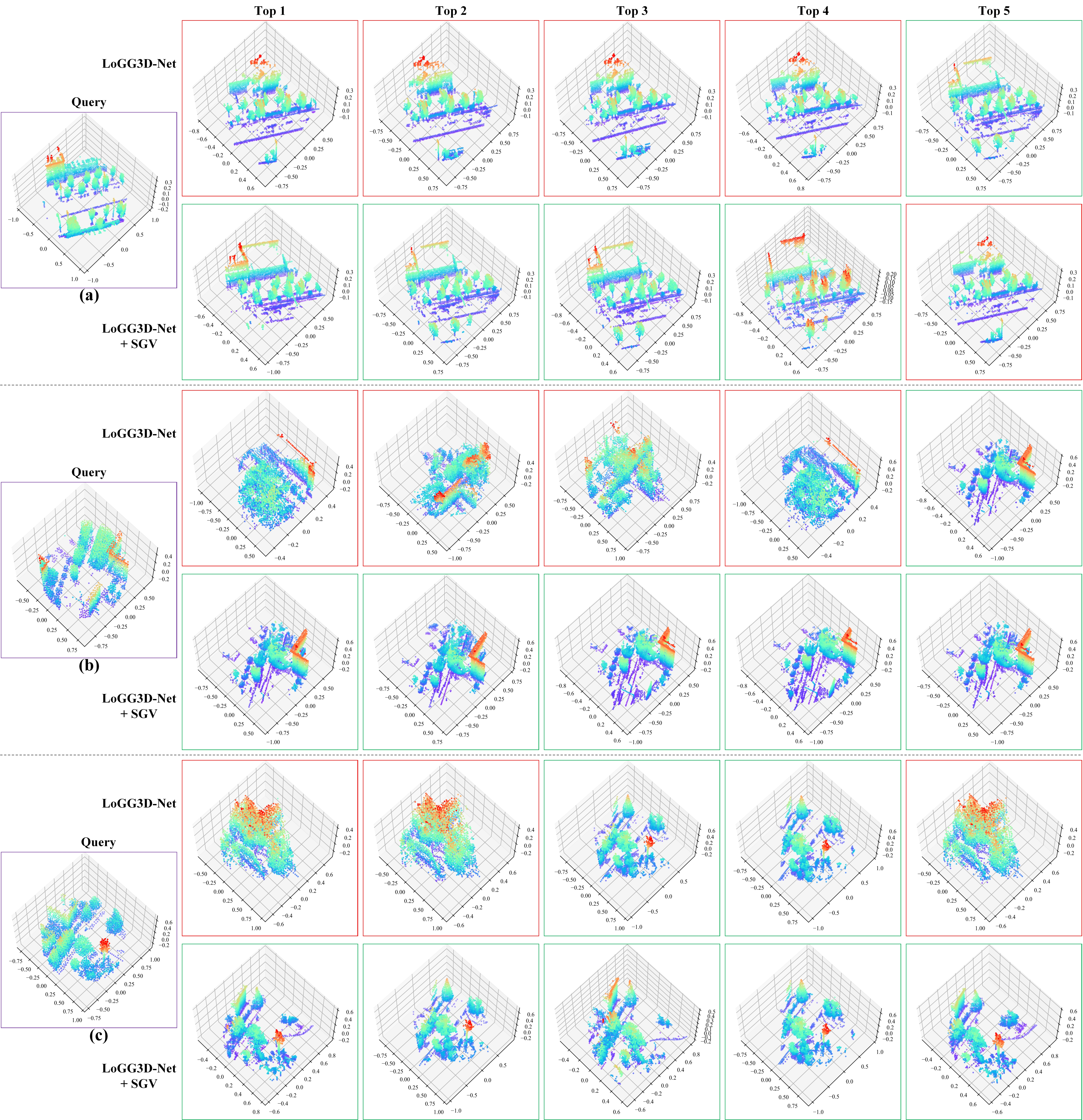}
    \caption{Success PR cases when using LoGG3D-Net for retrieval and SGV for reranking on WHU-PCPR with protocol 1. (a) case 1 on Hankou, (b) case 2 on WHU, (c) case 3 on WHU. Purple represents query, red represents failure, and green represents success.}
    \label{fig:success_cases}
\end{figure*}

Table \ref{tab:rerank_baselines_no_dynamics}, Table \ref{tab:rerank_baselines_with_dynamics}, and Fig. \ref{fig:success_cases} presents the results of reranking baselines. Overall, geometric consistency-based reranking method SGV achieves the best performance, while other reranking methods offer only marginal improvements. We analyze these evaluation results from the perspectives of the impacts of different retrieval methods and scenes, as well as computational efficiency.

\textbf{Impacts of different retrieval methods on reranking.}
As shown in Table \ref{tab:rerank_baselines_no_dynamics}, when PPTNet is used for initial retrieval, even the best reranking method SGV yields no performance improvement. In contrast, when superior retrieval methods such as EgoNN and LoGG3D-Net are employed for initial retrieval, SGV significantly enhances the PR performance, with improvements of up to 12.17\% in $R@1$, 8.49\% in $R@5$, and 7.55\% in $P@5$. The experimental results indicate that: 1) built upon high-performing retrieval methods, reranking methods based on geometry consistency can substantially enhance final performance; 2) the effectiveness of reranking is highly dependent on initial retrieval methods. This is because methods like EgoNN and LoGG3D-Net possess stronger feature extraction capabilities, producing more discriminative local features which are a prerequisite for the effectiveness of reranking methods based on geometric consistency.

\textbf{Impacts of different scenes and equipments on reranking.}
As shown in Table \ref{tab:rerank_baselines_no_dynamics}, when using LoGG3D-Net for initial retrieval and SGV for reranking, $R@1$ increases by 7.44\%, 10.12\%, 3.79\%, 3.69\%, and 12.17\% respectively on Hankou 1\&2, Hankou 1\&3, WHU 1\&2, WHU 1\&3, and WHU 2\&3, and $P@5$ increases by 7.23\%, 6.09\%, 1.49\%, 1.81\%, and 7.55\% respectively. The improvement in performance of SGV varies across different datasets, with the greatest improvement observed on WHU 2\&3, although it differs from the training set's scenes and acquisition equipments.

\textbf{Impacts of dynamic objects on reranking.}
As shown in \ref{tab:rerank_baselines_with_dynamics}, almost none of the re-ranking methods, including SGV, can improve place recognition performance under protocol 2. The main reason is that the submaps in protocol 2 contain a large number of dynamic objects, which pose great challenges to methods based on geometry verification like SGV.

\textbf{Efficiency.}
As shown in the last column of Table \ref{tab:rerank_baselines_no_dynamics_efficiency}, aQE achieves the highest computational efficiency, while RPR essentially achieves reranking by evaluating the results of point cloud registration, with the lowest efficiency. Meanwhile, the runtime of SGV remains consistently within 5 ms.

\textbf{PCPR cases.}
Fig. \ref{fig:success_cases} presents three successful cases of using LoGG3D-Net for retrieval and SGV for reranking. In these cases, the main scenes of query submaps are trees, with few artificial objects, causing LoGG3D-Net to return an incorrect top-1 candidate initially. SGV utilizes local features extracted by LoGG3D-Net and adjusts the initial retrieval results through spectral matching scores, effectively improving the PR performance.

\begin{figure*}
    \centering
    \includegraphics[width=1\linewidth]{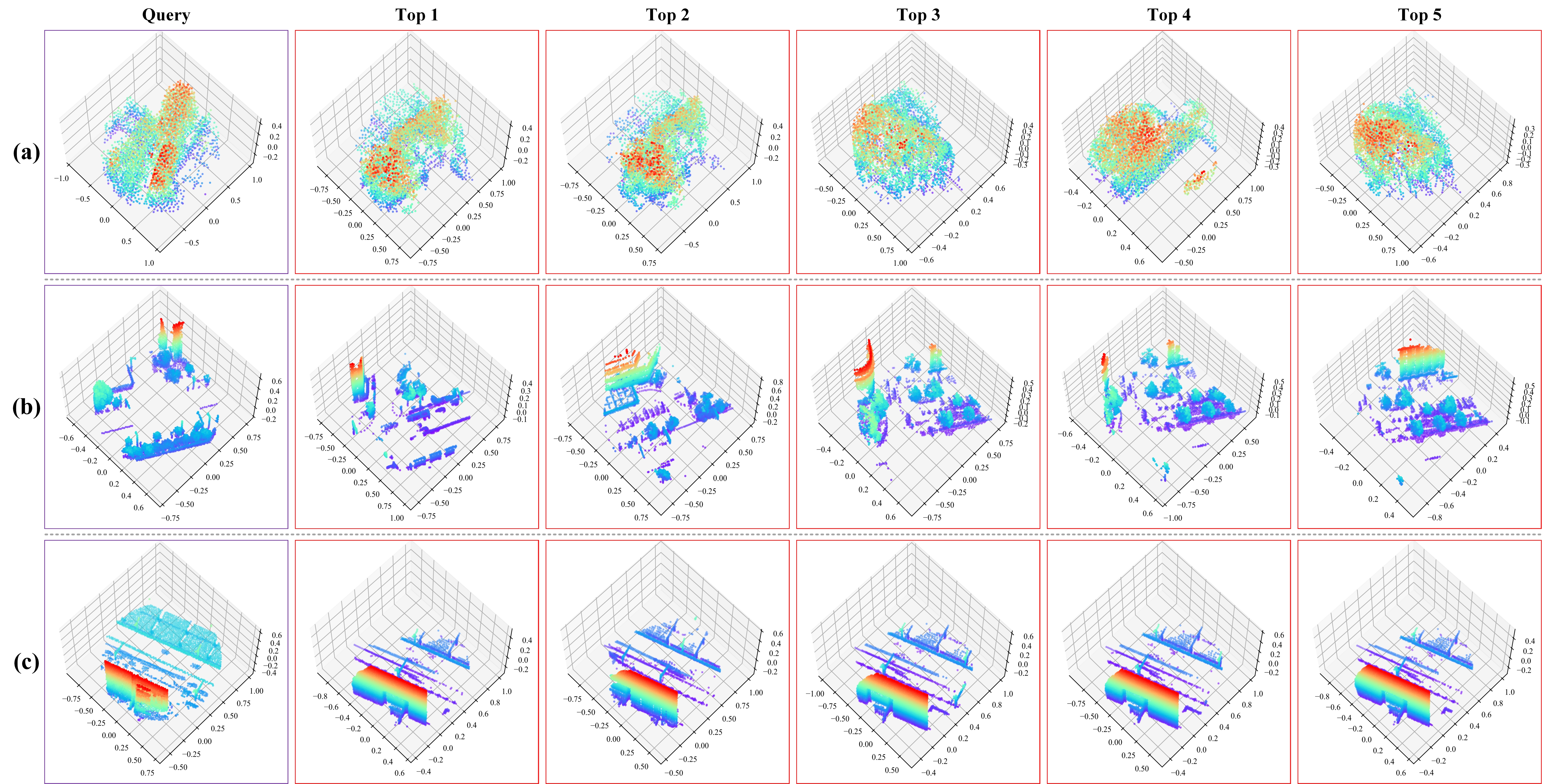}
    \caption{Failure cases when using LoGG3D-Net for retrieval and SGV for reranking on WHU-PCPR with protocol 1. (a) case 1, (b) case 2, (c) case 3.}
    \label{fig:Bad_cases}
\end{figure*}

Fig. \ref{fig:Bad_cases} presents failure cases using LoGG3D-Net and SGV. The three failure cases illustrate distinct challenges: Case 1 lacks sufficient features due to vegetation-only scenes, and such cases of perceptual degradation require the incorporation of sequential information. Case 2 is characterized by low distinctiveness at wide intersections, and such cases necessitate dynamic submap radius adjustment. Case 3 is misled by coincidental scene similarities in the map and has a higher probability of occurring in urban settings, necessitating the use of coarse localization technology. All are common corner cases where single-shot PCPR typically fails.

In summary, certain reranking methods can significantly enhance the PR performance on our challenging dataset, but they still face significant challenges posed by dynamic object interference. Among them, SGV, a reranking method based on geometric consistency, represents the current SOTA. However, research in this direction remains nascent, indicating substantial potential for future development.

\begin{table*}
\centering
\fontsize{6.4}{10}\selectfont
\caption{Reranking results on WHU-PCPR with protocol 1.}
\label{tab:rerank_baselines_no_dynamics}
\begin{tabular}{cccccccccccccc}
\hline
\multirow{2}{*}{\begin{tabular}[c]{@{}c@{}}Rerank\\method\end{tabular}} & \multirow{2}{*}{\begin{tabular}[c]{@{}c@{}}Retrieval\\method\end{tabular}} & \multicolumn{2}{c}{Hankou 1\&2} & \multicolumn{2}{c}{Hankou 1\&3}  & \multicolumn{2}{c}{Hankou 2\&3}  & \multicolumn{2}{c}{WHU 1\&2}    & \multicolumn{2}{c}{WHU 1\&3}     & \multicolumn{2}{c}{WHU 2\&3}      \\ 
\cline{3-14}
                                                                        &                                                                            & R@1            & R@5/P@5                      & R@1            & R@5/P@5                       & R@1            & R@5/P@5                       & R@1            & R@5/P@5                      & R@1            & R@5/P@5                       & R@1            & R@5/P@5                        \\ 
\hline
\multirow{3}{*}{None}                                                   & PPTNet                                                                     & 56.71          & 71.85/54.50                  & 18.81          & 29.67/18.00                   & 24.20          & 36.00/23.52                   & 23.94          & 35.60/23.11                  & 6.17           & 11.67/5.87                    & 15.46          & 26.96/14.87                    \\
                                                                        & LoGG3D-Net                                                                 & 80.70          & 91.82/76.65                  & \uline{36.25}  & \uline{52.93}/\uline{33.94}   & \textbf{46.55} & \textbf{62.60}/\textbf{44.06} & 40.13          & \textbf{58.50}/36.94         & 18.85          & \uline{32.45}/16.46           & 28.36          & \uline{47.53}/25.52            \\
                                                                        & EgoNN                                                                      & 71.43          & 82.65/69.57                  & 24.69          & 33.88/24.02                   & 32.50          & 42.86/31.50                   & 36.39          & 48.92/34.75                  & 11.82          & 21.19/12.19                   & 17.51          & 28.85/17.07                    \\ 
\hline
\multirow{3}{*}{aQE}                                                    & PPTNet                                                                     & 53.97          & 64.45/51.58                  & 16.77          & 23.03/16.44                   & 17.74          & 23.83/17.56                   & 18.03          & 23.53/17.84                  & 4.65           & 7.53/4.63                     & 14.56          & 22.81/14.23                    \\
                                                                        & LoGG3D-Net                                                                 & 72.79          & 82.72/71.55                  & 31.41          & 41.38/30.82                   & 31.92          & 40.54/31.39                   & 30.42          & 40.73/29.76                  & 14.25          & 21.99/13.94                   & 24.41          & 38.85/24.01                    \\
                                                                        & EgoNN                                                                      & 63.13          & 73.59/61.82                  & 21.78          & 27.44/21.64                   & 23.09          & 29.01/22.66                   & 26.58          & 32.41/25.80                  & 11.08          & 15.83/10.91                   & 16.39          & 24.35/16.42                    \\ 
\hline
\multirow{3}{*}{SGV}                                                    & PPTNet                                                                     & 51.44          & 78.51/47.76                  & 16.41          & 32.00/15.24                   & 18.78          & 31.73/17.23                   & 17.32          & 31.44/16.87                  & 4.93           & 12.63/4.73                    & 14.81          & 30.94/13.71                    \\
                                                                        & LoGG3D-Net                                                                 & \uline{88.14}  & \textbf{94.76}/\uline{83.88} & \textbf{46.37} & \textbf{57.92}/\textbf{40.03} & \uline{46.48}  & \uline{52.98}/\uline{41.19}   & \textbf{43.92} & \uline{54.48}/\textbf{38.43} & \textbf{22.54} & \textbf{34.23}/\textbf{18.27} & \textbf{40.53} & \textbf{56.02}/\textbf{33.07}  \\
                                                                        & EgoNN                                                                      & \textbf{89.33} & \uline{92.25}/\textbf{85.11} & 35.65          & 40.72/31.05                   & 36.87          & 40.04/32.39                   & \uline{42.85}  & 50.53/\uline{37.40}          & \uline{22.05}  & 28.78/\uline{17.48}           & \uline{32.58}  & 41.92/\uline{26.58}            \\ 
\hline
\multirow{3}{*}{RPR}                                                    & PPTNet                                                                     & 56.71          & 71.85/54.50                  & 17.31          & 27.37/16.55                   & 17.94          & 27.02/17.41                   & 19.83          & 29.31/19.14                  & 5.34           & 10.06/5.08                    & 14.87          & 25.93/14.30                    \\
                                                                        & LoGG3D-Net                                                                 & 78.86          & 91.85/73.62                  & 32.65          & 48.00/30.21                   & 33.56          & 45.48/30.96                   & 33.05          & 47.94/30.23                  & 16.23          & 28.46/14.48                   & 27.04          & 45.96/24.30                    \\
                                                                        & EgoNN                                                                      & 71.43          & 82.65/69.57                  & 22.71          & 31.22/22.07                   & 24.24          & 32.38/23.39                   & 30.02          & 40.14/28.66                  & 10.40          & 18.57/10.75                   & 16.86          & 27.75/16.41                    \\
\hline
\end{tabular}
\end{table*}

\begin{table*}
\centering
\fontsize{6.4}{10}\selectfont
\caption{Reranking results on WHU-PCPR with protocol 2.}
\label{tab:rerank_baselines_with_dynamics}
\begin{tabular}{cccccccccccccc} 
\hline
\multirow{2}{*}{\begin{tabular}[c]{@{}c@{}}Rerank\\method\end{tabular}} & \multirow{2}{*}{\begin{tabular}[c]{@{}c@{}}Retrieval\\method\end{tabular}} & \multicolumn{2}{c}{Hankou 1\&2}                 & \multicolumn{2}{c}{Hankou 1\&3}                  & \multicolumn{2}{c}{Hankou 2\&3}                  & \multicolumn{2}{c}{WHU 1\&2}                  & \multicolumn{2}{c}{WHU 1\&3}                  & \multicolumn{2}{c}{WHU 2\&3}                   \\
                               &                                   & R@1            & R@5/P@5                      & R@1            & R@5/P@5                       & R@1            & R@5/P@5                       & R@1            & R@5/P@5                       & R@1            & R@5/P@5                       & R@1            & R@5/P@5                        \\ 
\hline
\multirow{3}{*}{None}          & PPTNet                            & 54.05          & 74.16/52.57                  & 36.83          & 52.76/35.42                   & \uline{50.58}  & \uline{68.95}/\uline{48.74}   & 25.05          & 37.39/23.81                   & 20.65          & 30.62/19.67                   & 40.97          & 57.91/38.38                    \\
                               & LoGG3D-Net                        & \textbf{63.97} & \uline{78.90}/\textbf{61.10} & \textbf{45.20} & \textbf{58.63}/\textbf{43.20} & \textbf{64.89} & \textbf{77.76}/\textbf{62.10} & \textbf{43.27} & \textbf{56.08}/\textbf{40.83} & \textbf{41.64} & \textbf{53.32}/\textbf{38.89} & \textbf{59.26} & \textbf{73.62}/\textbf{56.07}  \\
                               & EgoNN                             & 48.45          & 62.98/43.34                  & 30.22          & 42.25/29.32                   & 49.59          & 64.66/47.69                   & 34.47          & \uline{50.78}/32.43           & 34.45          & 46.94/32.78                   & 38.70          & 54.36/35.66                    \\ 
\hline
\multirow{3}{*}{AlphaQE}       & PPTNet                            & 51.31          & 63.62/49.17                  & 33.88          & 43.20/33.24                   & 37.43          & 46.19/36.63                   & 19.33          & 24.69/18.90                   & 16.42          & 21.15/16.21                   & 35.70          & 48.27/35.19                    \\
                               & LoGG3D-Net                        & 55.04          & 69.58/53.61                  & 38.17          & 37.49/33.97                   & 43.36          & 51.14/42.22                   & 31.66          & 38.18/30.84                   & 30.67          & 37.36/30.28                   & 51.05          & 62.20/50.32                    \\
                               & EgoNN                             & 34.67          & 47.91/32.98                  & 27.28          & 33.97/26.61                   & 36.13          & 44.40/35.19                   & 25.88          & 34.79/24.54                   & 26.73          & 33.40/26.26                   & 30.24          & 42.18/28.82                    \\ 
\hline
\multirow{3}{*}{SGV}           & PPTNet                            & 54.05          & 78.30/31.27                  & 33.84          & 54.08/22.24                   & 37.90          & 54.82/23.00                   & 20.88          & 33.35/14.66                   & 18.25          & 31.07/13.51                   & 39.30          & 60.35/24.79                    \\
                               & LoGG3D-Net                        & \textbf{63.97} & \textbf{81.02}/32.20         & \uline{41.47}  & \uline{57.61}/22.76           & 48.02          & 59.26/23.83                   & 35.25          & 46.31/\uline{33.56}           & \uline{36.48}  & \uline{48.04}/22.28           & \uline{56.93}  & \uline{72.48}/31.33            \\
                               & EgoNN                             & 48.45          & 69.50/27.02                  & 27.81          & 42.20/17.51                   & 37.53          & 51.56/22.35                   & 28.01          & 44.49/17.62                   & 30.22          & 45.84/19.45                   & 37.14          & 54.66/20.44                    \\ 
\hline
\multirow{3}{*}{RPR}           & PPTNet                            & 54.02          & 74.16/52.57                  & 33.84          & 48.53/32.54                   & 37.90          & 51.74/36.44                   & 20.88          & 31.08/19.89                   & 18.25          & 26.98/17.41                   & 39.30          & 55.56/36.81                    \\
                               & LoGG3D-Net                        & 62.48          & 78.47/\uline{59.81}          & 40.94          & 53.55/\uline{39.05}           & 47.40          & 57.34/45.32                   & \uline{35.63}  & 47.85/21.64                   & 36.11          & 46.35/\uline{33.81}           & 55.79          & 70.38/\uline{52.98}            \\
                               & EgoNN                             & 48.45          & 62.98/43.34                  & 27.81          & 38.95/26.96                   & 37.53          & 49.16/35.98                   & 28.01          & 41.28/26.50                   & 30.22          & 41.02/28.75                   & 37.14          & 52.18/34.23                    \\
\hline
\end{tabular}
\end{table*}

\begin{table*}
\centering
\fontsize{6.4}{10}\selectfont
\caption{Efficiency of reranking baseline methods.}
\label{tab:rerank_baselines_no_dynamics_efficiency}
\begin{tabular}{cccccccccc}
\hline
Reranking method    & \multicolumn{3}{c}{aQE}                     & \multicolumn{3}{c}{SGV}      & \multicolumn{3}{c}{RPR}        \\
Retrieval Method & PPT-Net      & LoGG3D-Net    & EgoNN        & PPT-Net & LoGG3D-Net & EgoNN & PPT-Net & LoGG3D-Net & EgoNN   \\ 
\hline
time (ms)        & \uline{0.27} & \textbf{0.23} & \uline{0.27} & 5.32    & 2.88       & 3.82  & 290.09  & 64.81      & 106.39  \\
\hline
\end{tabular}
\end{table*}

\section{Challenges and research directions}\label{sec_challeng_future_work}
PCPR primarily encounters three major challenges:

\textbf{Generalization.}
During long-term operation, localization systems often necessitate continuous operation across diverse scenes using varied data acquisition platforms. Point clouds collected by different equipments in different scenes have significant differences in scene characteristics, coverage, scanning pattern, density, accuracy, and noise level, namely domain gaps. These gaps are particularly salient in WHU-PCPR (see Fig. \ref{fig:WHU-PCPR-overview} and \ref{fig:wuhan_pcpr_characteristics}). Although recent methods like PatchAugNet \citep{zou2023patchaugnet} and HeLiOS \citep{2025HeLiOS} attempt to improve adaptability to domain gaps through self-supervised auxiliary tasks and overlap-based data mining, the generalization of existing PCPR methods remains limited and cannot generalize well to different scenes and equipment. In the future, we can transfer prior knowledge from visual models \citep{2025ImLPR} and introduce domain adaptation and domain generalization \citep{2020vLPD} to PCPR tasks.

\textbf{Scene changes.}
Localization scenes typically evolve dynamically over time, including real-time changes from objects like pedestrians and vehicles, as well as long-term changes from objects like buildings, trees. WHU-PCPR contains abundant examples of scenes undergoing both types of changes (see Fig. \ref{fig:wuhan_pcpr_characteristics}). For real-time changes, methods like Octomap \citep{2013OctoMap} can often be employed to filter out dynamic objects. For long-term changes, retrieving the correct places becomes challenging when relying solely on the limited information contained in one point cloud,  especially if the scene alterations are substantial. Usually, the robustness of PR and localization can be improved by leveraging sequential information from sequentially collected point clouds \citep{ma2022seqot} or applying point cloud registration to identify retrieval failures and significant scene changes \citep{2022Predicting_Alignability}.

\textbf{Viewpoint variations.}
When a localization system passes through the same place multiple times, significant viewpoint variations may occur, presenting a formidable challenge to PCPR. Recent methods like RPR-Net \citep{2023RPR} and RIA-Net \citep{2024RIA} have enhanced robustness to viewpoint variations by augmenting low-level rotation-invariant features. Meanwhile, approaches like OverlapTransformer \citep{ma2022overlaptransformer} extract rotation-invariant features from range images. In summary, relevant research remains at an early stage, and WHU-PCPR can provide effective data support for such studies.

\textbf{Pose estimation.}
PR determines only a coarse location within the map, with accuracy limited by the density of map submaps. To enhance precision, it is often combined with pose estimation techniques such as point cloud registration. First, since PR utilizes global features while pose estimation relies on local features, extracting both simultaneously improves localization efficiency \citep{cattaneo2022lcdnet}. Second, Monte Carlo localization can be integrated to achieve robust localization by leveraging sequential point clouds. Key challenges lie in ensuring reliable initialization and detecting localization failures \citep{zou2025reliable}.

\section{Conclusion}\label{sec_conclusion}
This paper establishes WHU-PCPR, a cross-platform heterogeneous point cloud dataset for PR, which is collected in the urban area of Wuhan using high-precision MLS systems and portable PLS systems. The main features of this dataset are threefold: cross-platform heterogeneous point clouds, complex localization scenes, and large-scale spatial coverage.
Benchmark on WHU-PCPR reveals two main findings. Existing retrieval methods show considerable room for improvement in generalizing across domains and handling viewpoint variations. While reranking can effectively enhance initial retrieval results, research in this area remains limited. Future work should therefore focus on improving generalization and robustness to scene changes and viewpoint variations. By introducing WHU-PCPR, this paper aims to address the limitations of existing datasets and advance the frontier of PCPR tasks towards more complex and diverse urban environments, thereby enhancing the practicality of PCPR approaches.

\bibliography{ref}
\bibliographystyle{IEEEtranN}

\section*{Biography Section}
\begin{IEEEbiography}[{\includegraphics[width=1in,height=1.25in,clip,keepaspectratio]{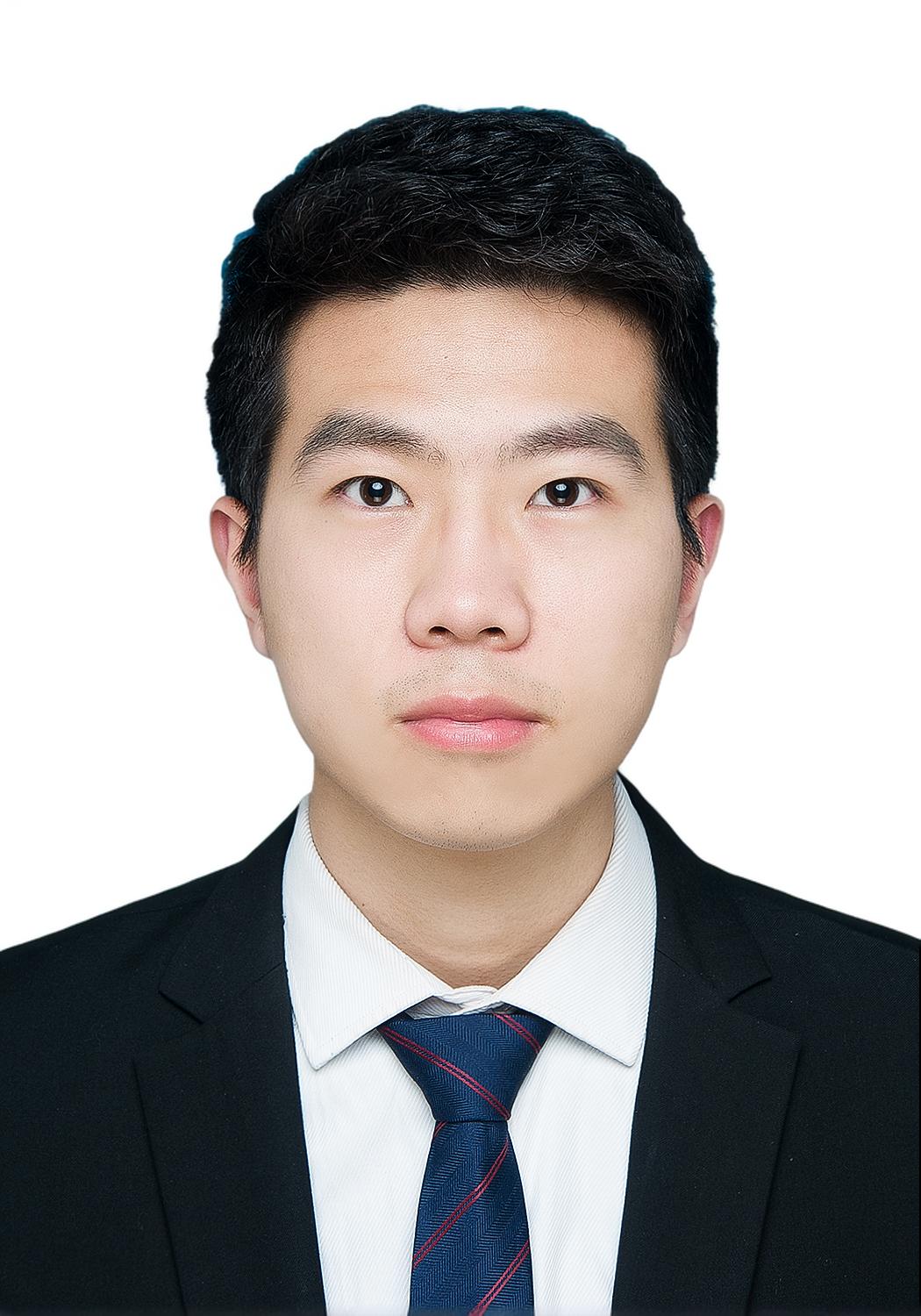}}]{Xianghong Zou}
received the B.S. degree in geomatics engineering, the M.S. degree, and the Ph.D. degree in photogrammetry and remote sensing from Wuhan University, Wuhan, China, in 2016, 2019, and 2024, respectively.

He is currently a post-doctoral researcher with the School of Advanced Manufacturing, Nanchang University, Nanchang, China. His research interests include point cloud data processing, global localization, and 3D change detection.
\end{IEEEbiography}

\begin{IEEEbiography}[{\includegraphics[width=1in,height=1.25in,clip,keepaspectratio]{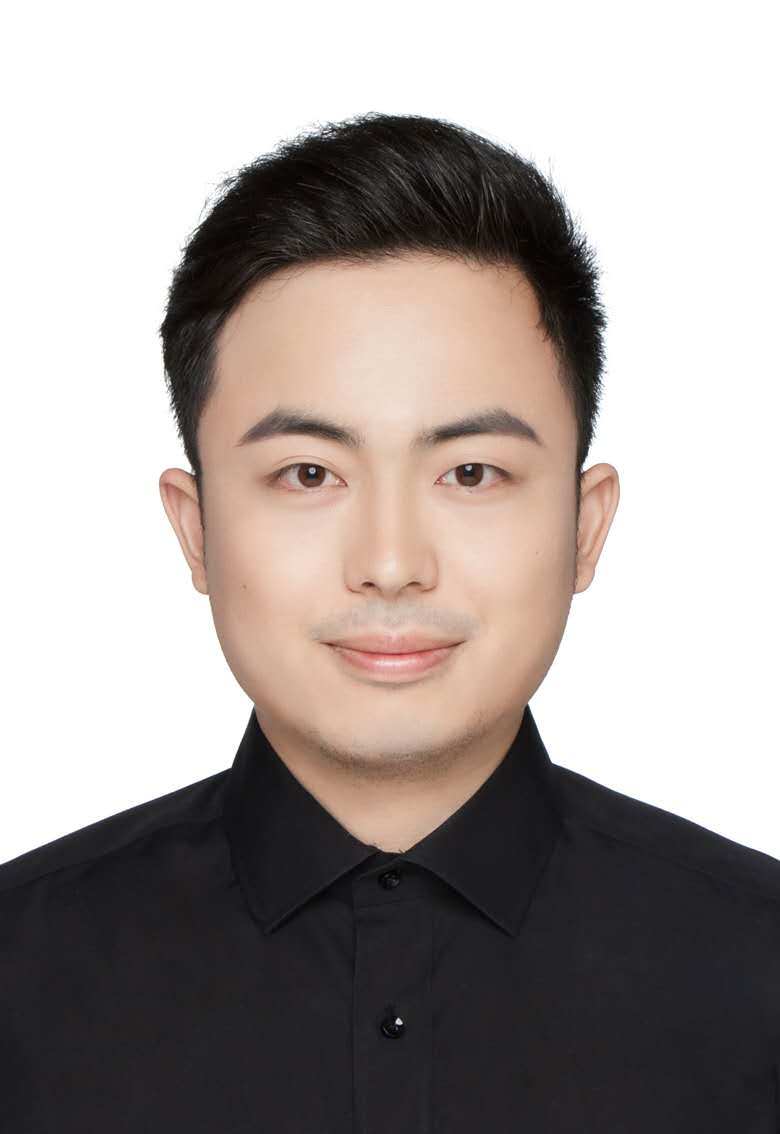}}]{Jianping Li}
received the B.S. degree in GIS and the Ph.D. degree in photogrammetry and remote sensing from Wuhan University, Wuhan, China, in 2015 and 2021, respectively. He is currently a Research Fellow with the School of Electrical and Electronic Engineering, Nanyang Technological University, Singapore. 

His research interests include 3D sensing system integration, UAV/UGV mapping, robot perception, and point cloud data processing.
\end{IEEEbiography}

\begin{IEEEbiography}[{\includegraphics[width=1in,height=1.25in,clip,keepaspectratio]{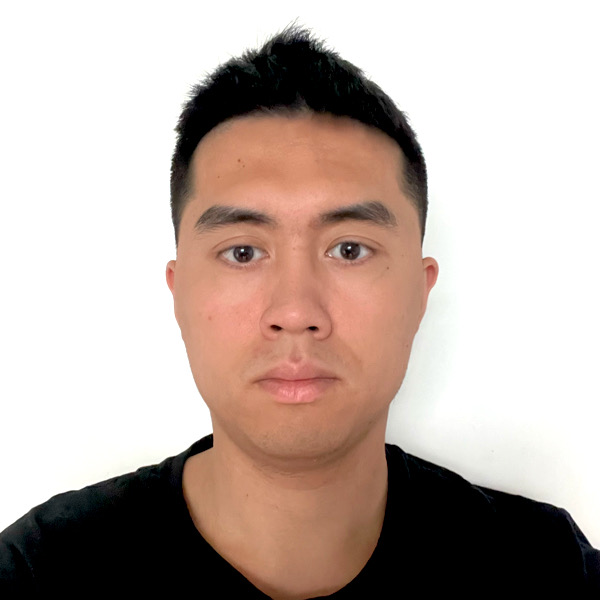}}]{Yandi Yang}
Yandi Yang received his M.E. degree in photogrammetry and remote sensing from the State Key Laboratory of Information Engineering in Surveying, Mapping and Remote Sensing, Wuhan University, Wuhan, China, in 2023. He is currently pursing his Ph.D. degree in geomatics engineering at the University of Calgary, Calgary, Candada.

His research interests include mobile mapping and sensor fusion.
\end{IEEEbiography}

\begin{IEEEbiography}[{\includegraphics[width=1in,height=1.25in,clip,keepaspectratio]{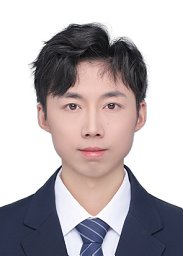}}]{Weitong Wu}
Weitong Wu received the B.E. degree in Geomatics Engineering from Central South University, Changsha, China, in 2017, and the Ph.D. degree in Photogrammetry and Remote Sensing from Wuhan University, Wuhan, China, in 2023. He is currently a lecturer with the School of Earth Sciences and Engineering, Hohai University, Nanjing, China.

His research interests include sensor calibration, SLAM, ubiquitous positioning, and 3D mapping.
\end{IEEEbiography}

\begin{IEEEbiography}[{\includegraphics[width=1in,height=1.25in,clip,keepaspectratio]{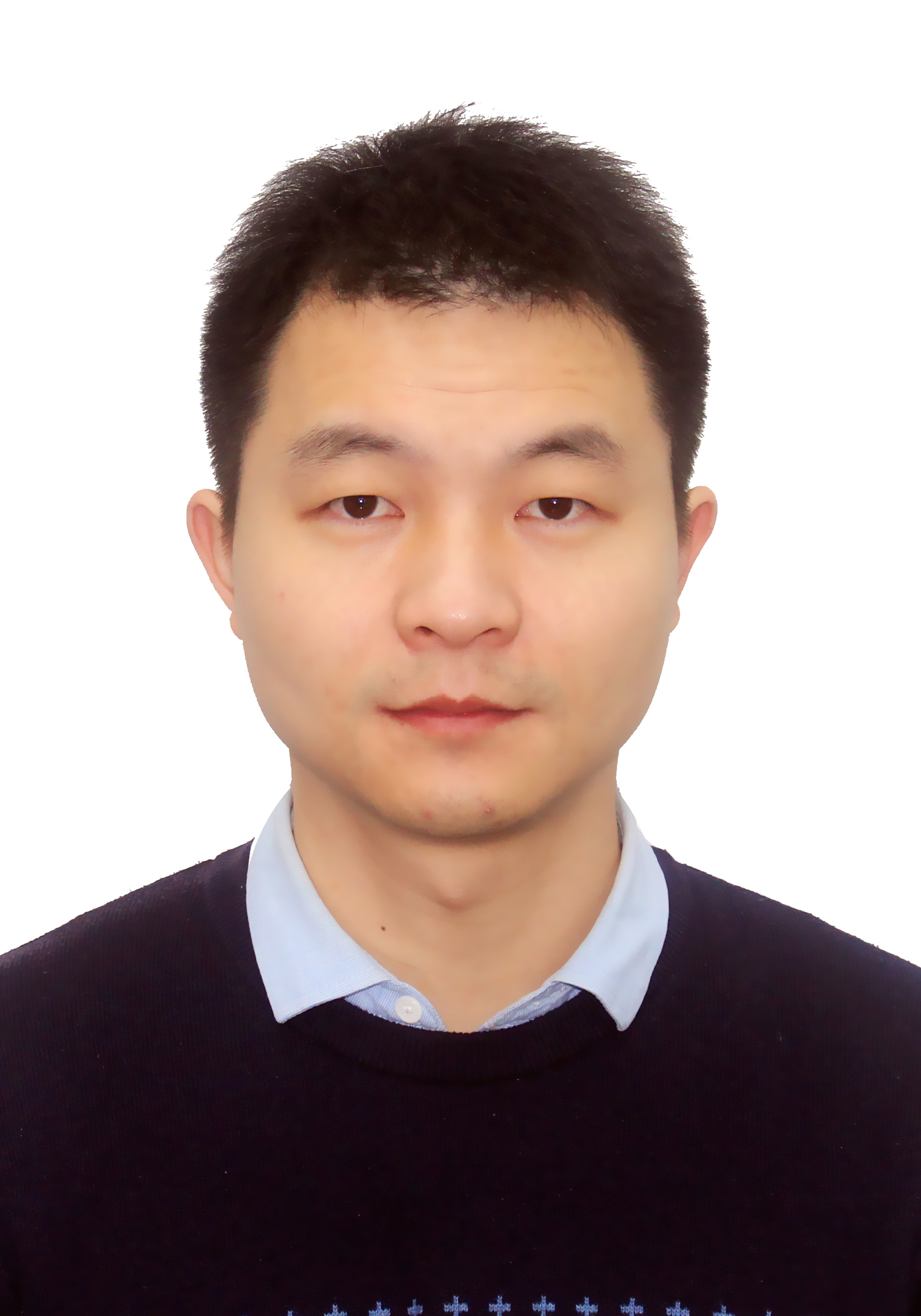}}]{Yuan Wang}
Yuan Wang received the B.S. degree in mathematics and applied mathematics from the Hefei University of Technology, Hefei, China, in 2013, and the Ph.D. degree from Wuhan University, Wuhan, China, in 2018. Since 2023, he has been with the School of Geography and Environment, Jiangxi Normal University, Nanchang, China.

His research interests include intelligent processing of point cloud data and multimodal data fusion.
\end{IEEEbiography}

\begin{IEEEbiography}[{\includegraphics[width=1in,height=1.25in,clip,keepaspectratio]{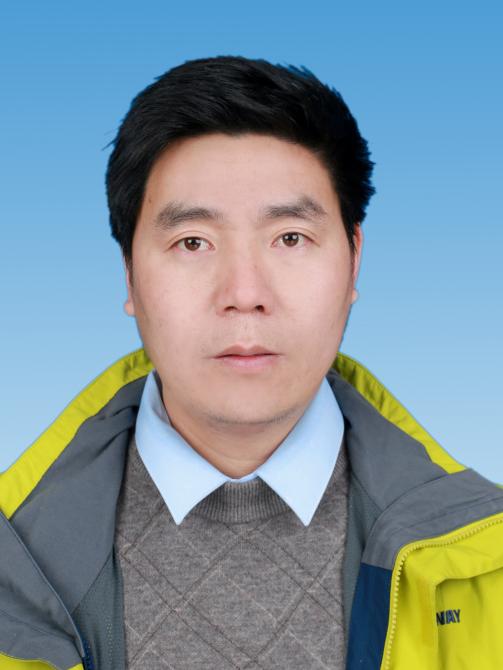}}]{Qiegen Liu}
received the Ph.D. degree in biomedical engineering from Shanghai Jiaotong University, Shanghai, China, in 2012.

During his time, he has been a Guest Speaker with the Institute of Computational Biology, Chinese Academy of Sciences (CAS), Beijing, China, and the Laubert Center for Medical Imaging, Shenzhen Institute of Advanced Technology (SIAT), Shenzhen, China. He did his Postdoctoral Work with UIUC, Champaign, IL, USA, in 2015, and the University of Calgary, Calgary, AB, Canada, in 2016. He is mainly engaged in sparse and deep learning representations and their applications in medical imaging and image processing.
\end{IEEEbiography}

\begin{IEEEbiography}[{\includegraphics[width=1in,height=1.25in,clip,keepaspectratio]{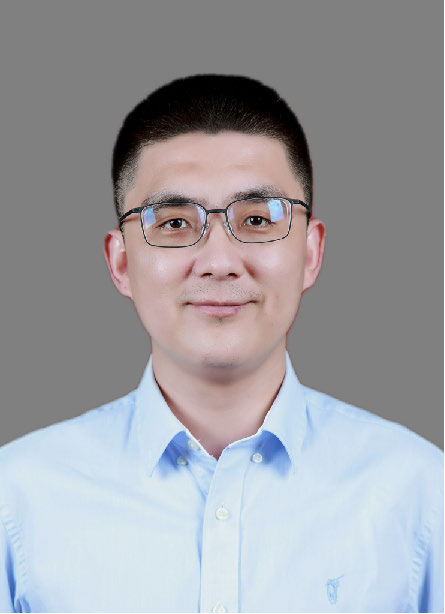}}]{Zhen Dong}
received the B.E. and Ph.D. degrees in remote sensing and photogrammetry from Wuhan University, in 2011 and 2018, respectively. He is currently a Professor of the State Key Laboratory of Information Engineering in Surveying, Mapping and Remote Sensing (LIESMARS), Wuhan University.

His research interest lies in the field of 3D Computer Vision, particularly including 3D reconstruction, scene understanding, point cloud processing as well as their applications in the intelligent transportation systems, digital twin cities, urban sustainable development and robotics.
\end{IEEEbiography}


\end{document}